\documentclass[pdflatex,sn-mathphys-num, iicol]{sn-jnl}

\usepackage{graphicx}%
\usepackage{multirow}%
\usepackage{amsmath,amssymb,amsfonts}%
\usepackage{amsthm}%
\usepackage{mathrsfs}%
\usepackage[title]{appendix}%
\usepackage{xcolor}%
\usepackage{textcomp}%
\usepackage{manyfoot}%
\usepackage{booktabs}%
\usepackage{algorithm}%
\usepackage{algorithmicx}%
\usepackage{algpseudocode}%
\usepackage{listings}%
\usepackage{hyperref}
\usepackage{pifont}
\usepackage[table,xcdraw]{xcolor}
\usepackage{longtable}
\usepackage{makecell}
\newcommand{\cmark}{\ding{51}} 
\newcommand{\xmark}{\ding{55}} 

\theoremstyle{thmstyleone}%
\theoremstyle{thmstyletwo}%

\theoremstyle{thmstylethree}%

\begin{document}

\title[Article Title]{LightOcc: Lightweight Spatial Embedding for Efficient Vision-based 3D Occupancy Prediction}


\author[1]{\fnm{Jinqing} \sur{Zhang}}\email{zhangjinqing@buaa.edu.cn}

\author[2]{\fnm{Yanan} \sur{Zhang}}\email{yananzhang@hfut.edu.cn}

\author[1]{\fnm{Wenrui} \sur{Cai}}\email{wenrui\_cai@buaa.edu.cn}
\author*[1,3]{\fnm{Qingjie} \sur{Liu}}\email{qingjie.liu@buaa.edu.cn}
\author[1,3]{\fnm{Yunhong} \sur{Wang}}\email{yhwang@buaa.edu.cn}

\affil[1]{\orgdiv{State Key Laboratory of Virtual Reality Technology and Systems}, \orgname{Beihang University}}
\affil[2]{\orgdiv{School of Computer Science and Information Engineering}, \orgname{Hefei University of Technology}}
\affil[3]{\orgdiv{Hangzhou Innovation Institute}, \orgname{Beihang University}}


\abstract{Occupancy prediction has garnered increasing attention in recent years for its comprehensive fine-grained environmental representation and strong generalization to open-set objects. Nevertheless, mainstream occupancy prediction methods employ cumbersome voxel features as the scene representation, incurring substantial overheads in both memory and computation. When comparing the occupancy distribution in each spatial dimension, we find that the information entropy of the height dimension is much lower than the other two dimensions that constitute the Bird's Eye View (BEV) plane, which indicates that the height distribution of occupancy is easier to learn and predict. Accordingly, we propose Lightweight Spatial Embedding that can represent complete height information in a more compact way than voxel features, thus significantly enhancing its deployability. First, Single-Channel Occupancy is sampled from the multi-view depth distributions, which is then processed by Spatial-to-Channel mechanism to extract Lightweight Spatial Embeddings of different views by 2D convolution. These embeddings will interact with each other through the Lightweight Cross-View Interaction module to obtain the Unified Embedding, which can directly supplement BEV features with height information. Furthermore, we extract Edge-aware Spatial Embedding and apply Geometric Supervision on Spatial Embeddings, aiming to enhance their capability to represent spatial information. We also propose BEV-CutMix, a feature-level data augmentation strategy, to increase the diversity of the driving scenes. We integrate these innovative components into a pure 2D convolutional model, namely LightOcc. Sufficient experimental results show that LightOcc achieves state-of-the-art performance on multiple benchmarks while demonstrating significant efficiency advantages.}




\maketitle

\section{Introduction}

With the gradual popularization of autonomous driving, the requirements for the perception capabilities of autonomous driving are increasing. Unlike traditional close-set 3D object detection~\cite{mao20223d, qian20223d, wang2023multi, ma2024vision} that only predicts the 3D bounding boxes of objects in limited categories, 3D occupancy prediction~\cite{zhang2024vision, xu2025survey, shi2023grid} partitions the 3D scene into grid cells and predicts semantic labels for each voxel. It enables a more fine-grained understanding of 3D scenes, stronger generalization to open-set objects, and greater robustness to occlusion and irregularly shaped objects, thereby enhancing high-level autonomous driving capabilities.

\begin{figure}[t]
    \centering
    \includegraphics[width=0.95\linewidth]{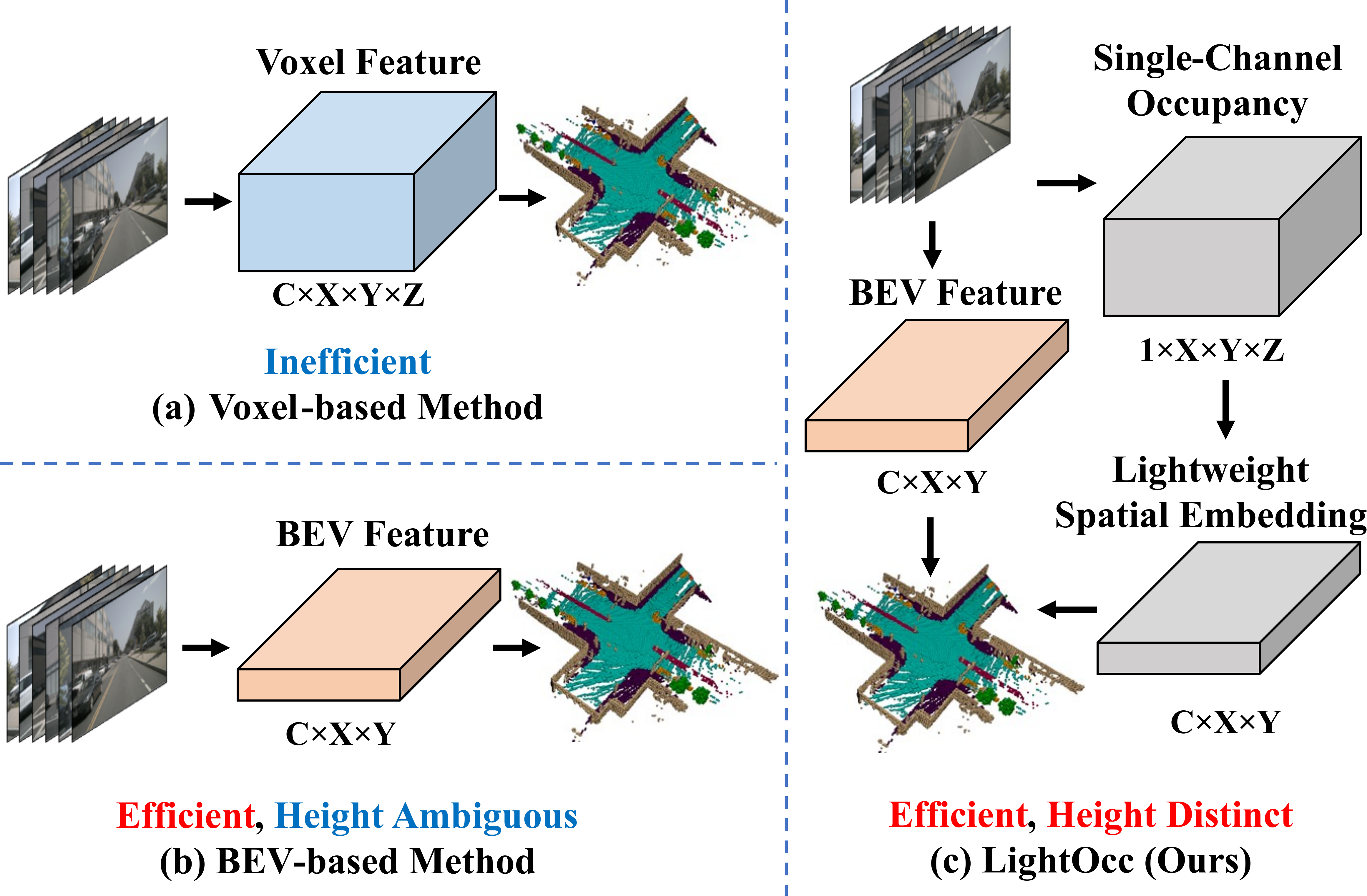}
    \caption{Comparison of different occupancy prediction frameworks. BEV-based methods replace voxel features with BEV features for better efficiency, but lose valuable height information. LightOcc extracts Lightweight Spatial Embedding that effectively supplements height information to BEV features, achieving higher accuracy without compromising efficiency.}
    \label{fig:compare}
\end{figure}

Since precise 3D occupancy prediction relies on understanding the overall 3D spatial information of driving scenes, current methods~\cite{li2023fbocc, hou2024fastocc, ma2024cotr, huang2021bevdet, wang2024panoocc, cao2022monoscene, tian2024occ3d} typically adopt voxel features as the representation as shown in Fig.~\ref{fig:compare}(a). This poses a significant challenge for practical deployment, as 3D voxel features consume too much memory and require time-consuming 3D convolutions or Transformer operations. To address the inefficiency of voxel-based approaches, some works try to construct sparse frameworks~\cite{liu2024fully, wang2024opus} or generate more compact representations~\cite{ma2024cotr,oh20253d}. However, we find that the three spatial dimensions of the driving scenes exhibit different amounts of information. Taking the Occ3D-nuScenes~\cite{tian2024occ3d} dataset as an example, the occupied voxels are evenly distributed along the X and Y dimensions that constitute the BEV plane, which has information entropy of $E_X=7.63$, $E_Y=7.38$. For Z dimension, most occupied voxels are concentrated at only a few height values around the ground, resulting in low information entropy of $E_Z=3.79$. It indicates that height information is relatively easy to learn and predict. On the other hand, the concentrated height distribution increases the sparsity of voxel features, hindering the capability of voxel-based methods to process height information. This explains why efficient BEV-based methods~\cite{yu2023flashocc,wu2024deep, duan2025sdgocc} shown in Fig.~\ref{fig:compare}(b) can also achieve comparable accuracy without modeling the height dimension.

Building on this insight, we decouple the cumbersome voxel features into BEV features and Lightweight Spatial Embedding, a compact representation of sparse height information, as illustrated in Fig.~\ref{fig:compare}(c). We first obtain the Single-Channel Occupancy by sampling the multi-view depth distributions originally utilized for BEV feature generation. Spatial-to-Channel Mechanism then takes each spatial dimension of Single-Channel Occupancy as the channel dimension, which allows the 2D convolutions to extract Lightweight Spatial Embeddings at different views. As shown in Fig.~\ref{fig:spatial2channel}, BEV Embedding is derived by taking Z dimension as the feature channel, thus implicitly representing height information. In contrast, Front View (FV) Embedding and Side View (SV) Embedding are obtained by treating X and Y dimensions as feature channels, storing local height information in the Z dimension that they retain. To obtain more complete height information, we introduce Lightweight Cross-View Interaction, which rapidly expands the receptive field of embedding elements to cover the entire driving scene through simple matrix multiplication. After iterative interactions, the embeddings of three views are fused into the Unified Embedding, providing rich height information to the BEV features. 

To enhance the sensitivity of Spatial Embeddings to object edges in the driving scene, we calculate the gradient of the Single-Channel Occupancy as the edge representation and extract Edge-aware Spatial Embeddings in the same processes as described above. We also apply Geometric Supervision to Spatial Embeddings to improve their ability to represent the entire spatial information. In addition, an augmentation strategy named BEV-CutMix is proposed to increase data diversity by cutting and mixing BEV features from different scenes.

Integrating the aforementioned components, we propose a novel 3D occupancy prediction framework, entirely built upon 2D convolutions, termed LightOcc. Experiments conducted on the multiple occupancy prediction benchmarks demonstrate that LightOcc achieves state-of-the-art results while exhibiting significant efficiency advantages. The major contributions of this paper can be summarized as:

\begin{itemize}
    \item  We decouple the original voxel representation of driving scenes into BEV features and Lightweight Spatial Embeddings based on the insight that the height information is much sparser and easier to learn.
    \item We implement the Lightweight Cross-View Interaction to effectively integrate both explicit and implicit height information carried by Lightweight Spatial Embeddings of different views into Unified Spatial Embeddings.
    \item We extract Edge-aware Spatial Embedding and apply Geometric Supervision to enhance the embeddings' ability to represent the entire spatial information. BEV-CutMix also effectively increases the data diversity.
    \item We conduct experiments on multiple occupancy prediction benchmarks, demonstrating the effectiveness and efficiency of Lightweight Spatial Embedding.
\end{itemize}

\section{Related Work}
\begin{figure*}[t]
    \centering
    \includegraphics[width=0.95\linewidth]{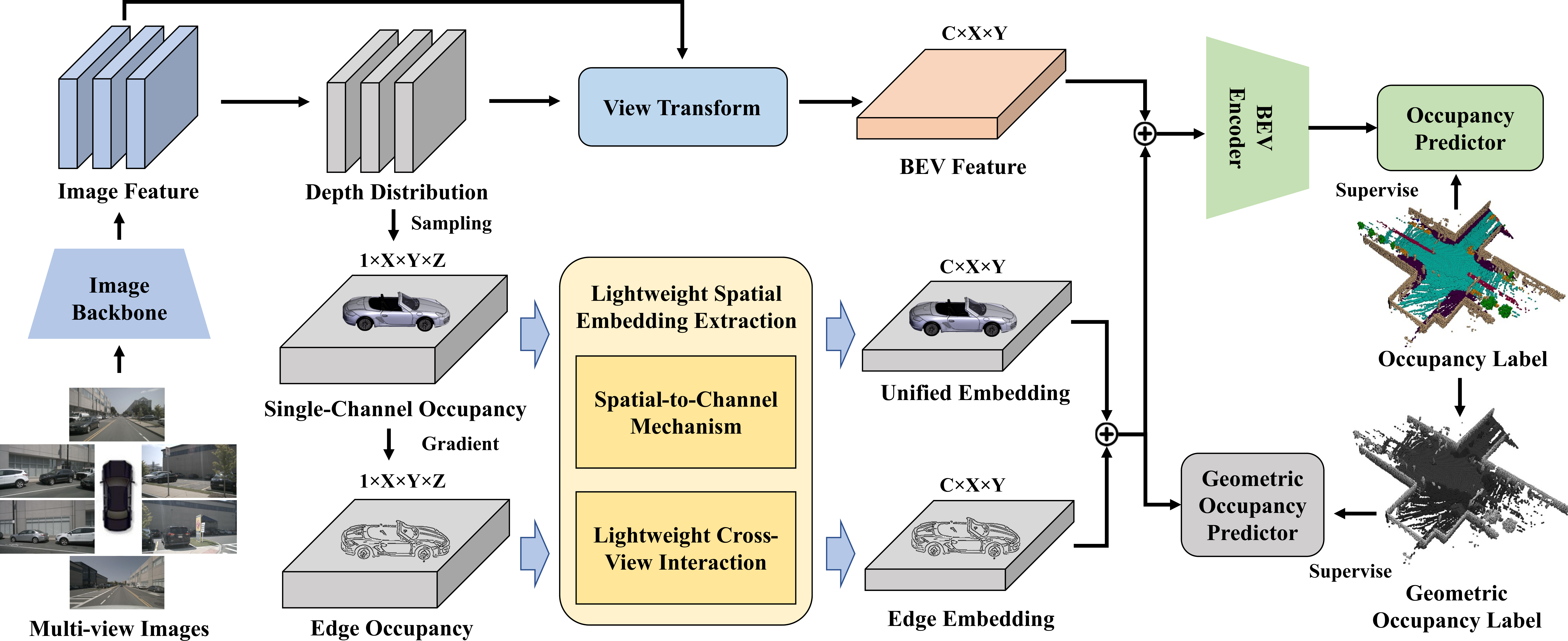}
    \caption{The overall framework of LightOcc. Single-Channel Occupancy is sampled from depth distributions, which is used to extract Lightweight Spatial Embeddings via Spatial-to-Channel mechanism and Lightweight Cross-View Interaction. Geometric Supervision is applied to enhance the embeddings' ability to represent complete spatial information.}
    \label{fig:lightocc}
\end{figure*}
\subsection{BEV Representation}

The BEV-based perception framework is effective for integrating multi-view image features into a uniform BEV representation, which improves performance in tasks like 3D object detection and BEV map segmentation. LSS~\cite{philion2020lift} weights the image features at different depths to generate pseudo-points, which are accumulated according to coordinates to obtain BEV representation. BEVDet~\cite{huang2021bevdet} constructs the BEV-based 3D object detection framework and proposes augmentation strategies in BEV space. BEVDet4D~\cite{huang2022bevdet4d} fuses the BEV features from past frames to help predict the velocity of the objects. BEVDepth~\cite{li2023bevdepth} and BEVStereo~\cite{li2023bevstereo} introduce the camera parameters and multi-view stereo to optimize the depth prediction. SA-BEV~\cite{zhang2023sa} further integrates the semantic information from the perspective view into the BEV representation. FSD-BEV~\cite{jiang2025fsd} proposes a self-distillation framework of BEV features. BEVNext~\cite{li2024bevnext} utilizes modules with extended receptive fields to aggregate long-term temporal information. GeoBEV~\cite{zhang2024geobev} proposes RC-Sampling to efficiently generate BEV features with high geometric quality.

Another branch of BEV-based perception methods uses Transformer to adaptively retrieve the corresponding image features to generate BEV representation. BEVFormer~\cite{li2022bevformer} applies cross-attention and self-attention to aggregate the image features and previous BEV features. BEVFormerV2~\cite{yang2023bevformer} adds a 2D detection head and applies 2D detection supervision to optimize the image features. PolarFormer~\cite{jiang2023polarformer} transforms the BEV space into the polar space and designs a special detection head. DFA3D~\cite{li2023dfa3d} inserts the explicit depth distribution into the BEV features when doing the cross-attention and simplifies the calculation of 3D Transformer.

\subsection{Occupancy Prediction}

Compared to 3D object detection, occupancy prediction provides a more detailed and comprehensive representation of the driving scene, aiding the vehicle in making more reasonable decisions. It also requires the ability of BEV framework to aggregate multi-view images, but generally represents the scene with 3D voxel features~\cite{li2023voxformer,zhang2023occformer, cao2022monoscene}. The classic 3D object detectors such as BEVDet and BEVFormer can be modified into the occupancy prediction method by transforming the image features into voxel features. SurroundOcc~\cite{wei2023surroundocc} expands the height dimension of BEV features by adopting spatial cross-attention and applies 3D convolution to upsample the volume features. FB-OCC~\cite{li2023fbocc} combines forward view transformation with the backward approach to improve the BEV features, which are combined with voxel features for further processing. COTR~\cite{ma2024cotr} reconstructs the compact 3D occupancy representation that has higher information density than voxel features. OccMamba~\cite{li2025occmamba} constructs the first Mamba framework for occupancy prediction.
Some methods pursue annotation-free training. RenderOcc~\cite{pan2024renderocc} predicts the density and label for each voxel based on the voxel features and renders the depth map and semantic segmentation of the images. SelfOcc~\cite{huang2024selfocc} employs a signed distance field to render 2D depth maps, along with raw color and semantic maps, under the supervision of the image sequence.

Unlike the approaches that utilize voxel features to represent the information of the driving scenes, several methods seek more efficient processing. TPVFormer~\cite{huang2023tri} divides 3D space into three perspective views and utilizes the attention mechanism to extract TPV features. S2TPVFormer~\cite{silva2024s2tpvformer} proposes Temporal Cross-View Hybrid Attention to utilize the history TPV features as the latent spatial representation. FastOcc~\cite{hou2024fastocc} replaces the voxel encoder with the BEV encoder to process BEV features and sample the image features to compensate for missing height information. FlashOcc~\cite{yu2023flashocc} predicts 3D occupancy only based on the BEV features to achieve the best efficiency. It argues that occupancies at different heights are closely related and can be predicted utilizing the same features. CVT-Occ~\cite{ye2024cvt} built 3D cost volume from temporally BEV features across historical frames. DHD~\cite{wu2024deep} follows the same concept and constructs three BEV features that represent the scene at different heights. After using extra modules to fuse these BEV features, performance improvement is achieved. OPUS~\cite{wang2024opus} represents the 3D scene as a sparse set of occupied points predicted directly by learnable queries, avoiding dense voxel grid modeling. Different from these approaches that discard height information or utilize heavy modules for compensation, we propose the Lightweight Spatial Embedding that preserves the comprehensive height information while maintaining real-time performance.

\section{Method}

\subsection{Preliminary of BEV-Based Occupancy Prediction}

Compared with voxel-based occupancy prediction methods, BEV-based methods such as FlashOcc~\cite{yu2023flashocc} exhibit significant advantages in perception efficiency and have high deployment potential in practical autonomous driving systems. We build FlashOcc-opt, an optimized version of FlashOCC, as the baseline of our method. It first utilizes a shared image backbone to extract image features $\{\textbf{F}_{I}^1, \textbf{F}_{I}^2,\cdots\textbf{F}_{I}^N\}$ from multi-view image$\{\textbf{I}_1, \textbf{I}_2,\cdots\textbf{I}_N\}$.
After that, the image features $\textbf{F}_{I}^i$ of each view are processed by the DepthNet module to predict the depth distribution $\textbf{D}_{i}$ with the same spatial shape as $\textbf{F}_{I}^i$. The channel dimension $\textbf{D}_{i}$ is D, representing the number of discrete depth values. Each element in $textbf{D}_{i}$ indicates the likelihood of the corresponding image feature at a specific depth value. Meanwhile, $\textbf{F}_{I}^i$ are also transformed into contextual features $\textbf{F}_{C}^i$ via a convolutional layer, 
which is formulated as:
\begin{equation}
    \begin{aligned}
    \textbf{D}_{i} &= {\rm DepthNet}(\textbf{F}_{I}^i) \\
    \textbf{F}_{C}^i &= {\rm Conv}(\textbf{F}_{I}^i)
    \end{aligned}
\end{equation}
For the lightweight model, DepthNet can be implemented with a single convolutional layer, while for higher-accuracy models, the DepthNet module designed by BEVDepth or BEVStereo can be adopted and supervised using depth labels generated from LiDAR point clouds.

Subsequently, the model employs the RCSampling module proposed by GeoBEV~\cite{zhang2024geobev} to convert depth distribution and contextual features of multiple views into unified BEV features with high geometry quality, which is formulated as:
\begin{equation}
    \textbf{F}_{BEV} = \sum_i^N{\rm RC\mbox{-}Sampling}(\textbf{F}_{C}^i,\textbf{D}_{i}).
\end{equation}

After further feature extraction of BEV features by the BEV encoder, the model adopts the channel-to-height mechanism proposed by FlashOcc to directly predict 3D occupancy $\mathbf{O}^{Pred}$ using 2D BEV features, which can be represented by:
\begin{equation}
    \mathbf{O}^{Pred}={\rm OccHead}(\textbf{F}_{BEV})\rightarrow \mathbb{R}^{\scriptscriptstyle S\times X\times Y\times Z}
\end{equation}
where OccHead consists of 2D convolutional layers and $S, X, Y, Z$ denote the semantic category and the length, width, height of the occupancy, respectively.

\subsection{Overall Architecture}

The overall framework of our proposed LightOcc is shown in Fig.~\ref{fig:lightocc}. Following the extraction of image features from multi-view images, the depth distribution is predicted to facilitate the view transformation to the BEV space. Single-Channel Occupancy is then sampled to aggregate the spatial information across views. Spatial-to-Channel mechanism enables the model to extract Lightweight Spatial Embeddings using 2D convolutions. These embeddings are fused through Lightweight Cross-View Interaction, producing Unified Embedding that enriches BEV features with comprehensive spatial information. Additionally, Edge-aware Spatial Embeddings are extracted to better represent edge information. Geometric Occupancy is used to supervise the geometry occupancy predicted by Unified Embedding and Edge Embedding to improve their quality. To more intuitively demonstrate the pipeline of LightOcc, we present it in the form of pseudo-code in Alg.~\ref{alg:lightocc}.

\begin{algorithm}[t]
\caption{Pseudocode pipeline of LightOcc}
\label{alg:lightocc}
\renewcommand{\algorithmicrequire}{\textbf{Input:}}
\renewcommand{\algorithmicensure}{\textbf{Output:}}
\begin{algorithmic}
\Require{Multi-view images $\{\mathbf{I}_1, \mathbf{I}_2, \dots, \mathbf{I}_N\}$}
\Ensure{3D occupancy prediction $\mathbf{O}^{Pred} $}

\For{$i = 1$ \textbf{to} $N$}
    \State $\mathbf{F}_I^i \leftarrow \text{Backbone}(\mathbf{I}_i)$\Comment{Image features}
    \State $\mathbf{D}_i \leftarrow \text{DepthNet}(\mathbf{F}_I^i)$\Comment{Depth distribution}
    \State $\mathbf{F}_C^i \leftarrow \text{Conv}(\mathbf{F}_I^i)$ \Comment{Context features}
\EndFor

\Statex $\triangleright$ Obtain BEV features
\State $\mathbf{F}_{BEV} \leftarrow \sum_{i=1}^N \text{RC-Sampling}(\mathbf{F}_C^i, \mathbf{D}_i)$ 

\Statex $\triangleright$ Obtain Single-Channel Occupancy
\State $\mathbf{O}_{SC} \leftarrow \text{DepthSampling}(\{\mathbf{D}_1, \mathbf{D}_2, \dots, \mathbf{D}_N\})$

\Statex $\triangleright$ Obtain Edgea-aware Occupancy
\State $\mathbf{O}_{SC}^{Edge} \leftarrow \text{SpatialGradient}(\mathbf{O}_{SC})$ 

\Function{EmbeddingExtraction}{$\mathbf{O}_{SC}$}
    \State $\triangleright$ Spatial-to-Channel Mechanism
    \State $\{\mathbf{E}_{BEV},\mathbf{E}_{FV}, \mathbf{E}_{SV}\} \leftarrow \text{S2C}(\mathbf{O}_{SC})$ 
    \State $\triangleright$ Lightweight Cross-View Interaction
    \State $\mathbf{E}_{U} \leftarrow \text{LCVI}(\{\mathbf{E}_{BEV},\mathbf{E}_{FV}, \mathbf{E}_{SV}\})$
    \State \Return $\mathbf{E}_{U}$ 
\EndFunction

\Statex $\triangleright$ Obtain Unified Spatial Embeddings
\State $\mathbf{E}_{U} \leftarrow \textsc{EmbeddingExtraction}(\mathbf{O}_{SC})$ 
\Statex $\triangleright$ Obtain Edge-Aware Spatial Embeddings
\State $\mathbf{E}_U^{Edge} \leftarrow \textsc{EmbeddingExtraction}(\mathbf{O}_{SC}^{Edge})$
\State $\mathbf{F}_{BEV}^{\prime} \leftarrow \mathbf{F}_{BEV} + \mathbf{E}_{U} + \mathbf{E}_U^{Edge}$

\State $\mathbf{F}_{BEV}^{\prime\prime} \leftarrow \text{BEVEncoder}(\mathbf{F}_{BEV}^{\prime})$
\Statex $\triangleright$ Channel-to-Height Mechanism
\State $\mathbf{O}^{Pred} \leftarrow \text{C2H}(\mathbf{F}_{BEV}^{\prime\prime})$ 

\If{isTraining}
    \State $\triangleright$ Geometric supervision
    \State $\mathbf{O}_{geo}^{Pred} \leftarrow \text{C2H}(\mathbf{E}_{U}+\mathbf{E}_U^{Edge})$
    \State $\mathcal{L}_{main} \leftarrow \text{MainOccLoss}(\mathbf{O}^{Pred}, \mathbf{O}^{GT})$ 
    \State $\mathcal{L}_{geo} \leftarrow \text{GeoOccLoss}(\mathbf{O}_{geo}^{Pred}, \mathbf{O}_{geo}^{GT})$ 
    \State $\mathcal{L}_{total} \leftarrow \mathcal{L}_{occ} + \cdot \mathcal{L}_{geo}$
\EndIf

\State \Return $\mathbf{O}^{Pred}$
\end{algorithmic}
\end{algorithm}

\subsection{Single-Channel Occupancy}

Recent improvements~\cite{huang2022bevpoolv2, zhang2024geobev} to LSS~\cite{philion2020lift} have substantially boosted the efficiency of BEV feature generation by obviating the requirement for heavy 3D intermediate features. However, this results in the loss of height information stored within 3D intermediate features. Consequently, retaining height information without depending on 3D intermediate features remains a pivotal challenge for advancing BEV-based occupancy prediction approaches.

Given that the information entropy in the height dimension is significantly smaller than that in the BEV plane, we assume that single-channel 3D features are sufficient to represent the information lost by BEV features. Consequently, we sample the multi-view depth distributions used for view transformation to generate a Single-Channel Occupancy. Defining $D$ as the predicted depth channel and $H, W$ as the height and width of image features, the depth distribution can be represented as $\mathbf{D}\in \mathbb{R}^{\scriptscriptstyle D\times H\times W}$. We pre-define the $(x,y,z)$ as the 3D coordinates of Single-Channel Occupancy $\mathbf{O}_{SC}\in \mathbb{R}^{\scriptscriptstyle 1\times X\times Y \times Z}$, where $X, Y, Z$ are the length, width and height of the 3D space, and project them into the image space as follows:
\begin{equation}
    d(h,w,1)^{\rm T} = \textbf{K}\left[\textbf{R}(x,y,z)^{\rm T}+\textbf{t}\right]
\end{equation}
where $\textbf{R}, \textbf{t}, \textbf{K}$ represent the rotation matrix, translation vector and intrinsic matrix of the camera. Regarding $(d,h,w)$ as the coordinates in $\mathbf{D}$,  $\mathbf{O}_{SC}$ can be formulated below:
\begin{equation}
    \mathbf{O}_{SC}(x,y,z) = \sum_{i=1}^{N}{\rm Sampling}(\mathbf{D}_{i}, (d_i, h_i, w_i))
\end{equation}
where $N$ is the number of multi-view images. 

Assuming that the original 3D intermediate features have $C$ channels, replacing them with $\mathbf{O}_{SC}$ to represent height information incurs only $\frac{1}{C}$ of the sampling time and memory usage, thereby preserving the high efficiency and strong deployability of the BEV-based method.

\begin{figure}[t]
    \centering
    \includegraphics[width=0.9\linewidth]{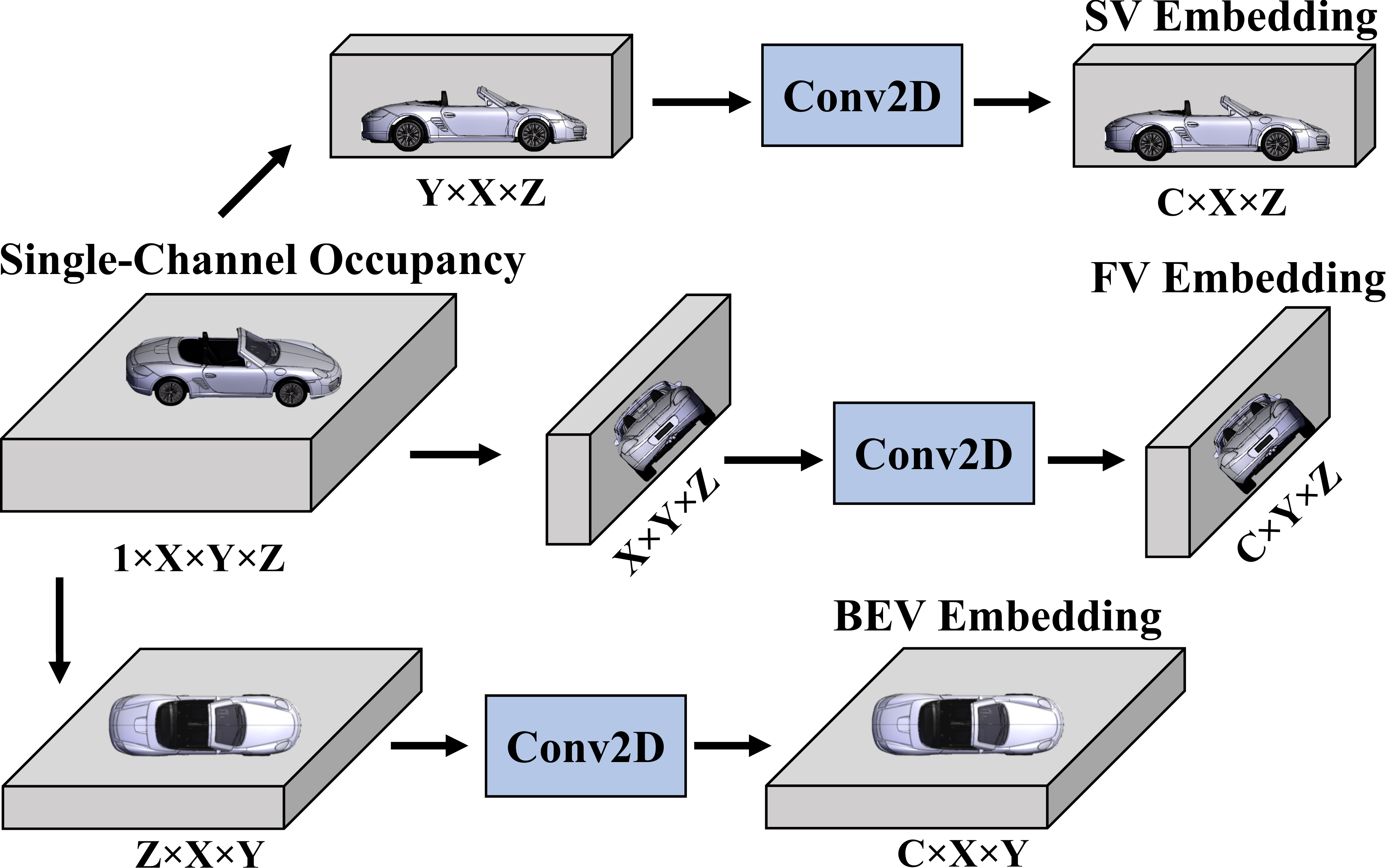}
    \caption{Illustration of Spatial-to-Channel Mechanism. Each spatial dimension of Single-Channel Occupancy can be treated as the feature channel.}
    \label{fig:spatial2channel}
\end{figure}

\subsection{Spatial-to-Channel Mechanism}

After obtaining $\mathbf{O}_{SC}$, effectively extracting height information and converting it into Spatial Embeddings remains a challenge. Unlike common approaches that process the $\mathbf{O}_{SC}$ using multiple 3D convolution layers, we propose the Spatial-to-Channel Mechanism, which treats each spatial dimension of $\mathbf{O}_{SC}$ as the feature channel. As shown in Fig.~\ref{fig:spatial2channel}, taking the Z (height) dimension as the feature channel, the BEV Embedding $\mathbf{E}_{BEV}\in \mathbb{R}^{\scriptscriptstyle C\times X\times Y}$ can be generated by:
\begin{equation}
    \mathbf{E}_{BEV} = {\rm Conv2D}(\mathbf{O}_{SC}\rightarrow \mathbb{R}^{\scriptscriptstyle Z\times X\times Y})
\end{equation}
where $\rightarrow$ indicates the matrix transpose operation.

Similarly, the Front View (FV) Embedding $\mathbf{E}_{FV}\in \mathbb{R}^{\scriptscriptstyle C\times Y\times Z}$ and Side View (SV) Embedding $\mathbf{E}_{SV}\in \mathbb{R}^{\scriptscriptstyle C\times X\times Z}$ can be obtained by:
\begin{align}
    \mathbf{E}_{FV} &= {\rm Conv2D}(\mathbf{O}_{SC}\rightarrow \mathbb{R}^{\scriptscriptstyle X\times Y\times Z}) \\
    \mathbf{E}_{SV} &= {\rm Conv2D}(\mathbf{O}_{SC}\rightarrow \mathbb{R}^{\scriptscriptstyle Y\times X\times Z})
\end{align}
Different from $\mathbf{E}_{BEV}$, which encodes implicit height information within its feature channels, $\mathbf{E}_{FV}$ and $\mathbf{E}_{SV}$ retain the spatial dimension of height, enabling explicit preservation of local height information.

For a single 3D convolution layer with a kernel size of 3, it can expand the features' receptive field to $3\times 3 \times 3$. On the contrary, Spatial-to-Channel Mechanism expands the receptive field of elements in $\mathbf{E}_{BEV}$ to $3\times 3 \times Z$. Experiments show that Spatial-to-Channel Mechanism is more effective in extracting features from a low-entropy representation.

\subsection{Lightweight Cross-View Interaction}

While $\mathbf{E}_{BEV}$ shares the same spatial dimensions as the BEV features $\mathbf{F}_{BEV}\in \mathbb{R}^{\scriptscriptstyle C\times X\times Y}$, $\mathbf{E}_{FV}$ and $\mathbf{E}_{SV}$ cannot be directly fused with $\mathbf{F}_{BEV}$. Given the importance of the explicit height information carried by $\mathbf{E}_{FV}$ and $\mathbf{E}_{SV}$, it is essential to facilitate interaction between them and $\mathbf{E}_{BEV}$.

\begin{figure}[t]
    \centering
    \includegraphics[width=0.9\linewidth]{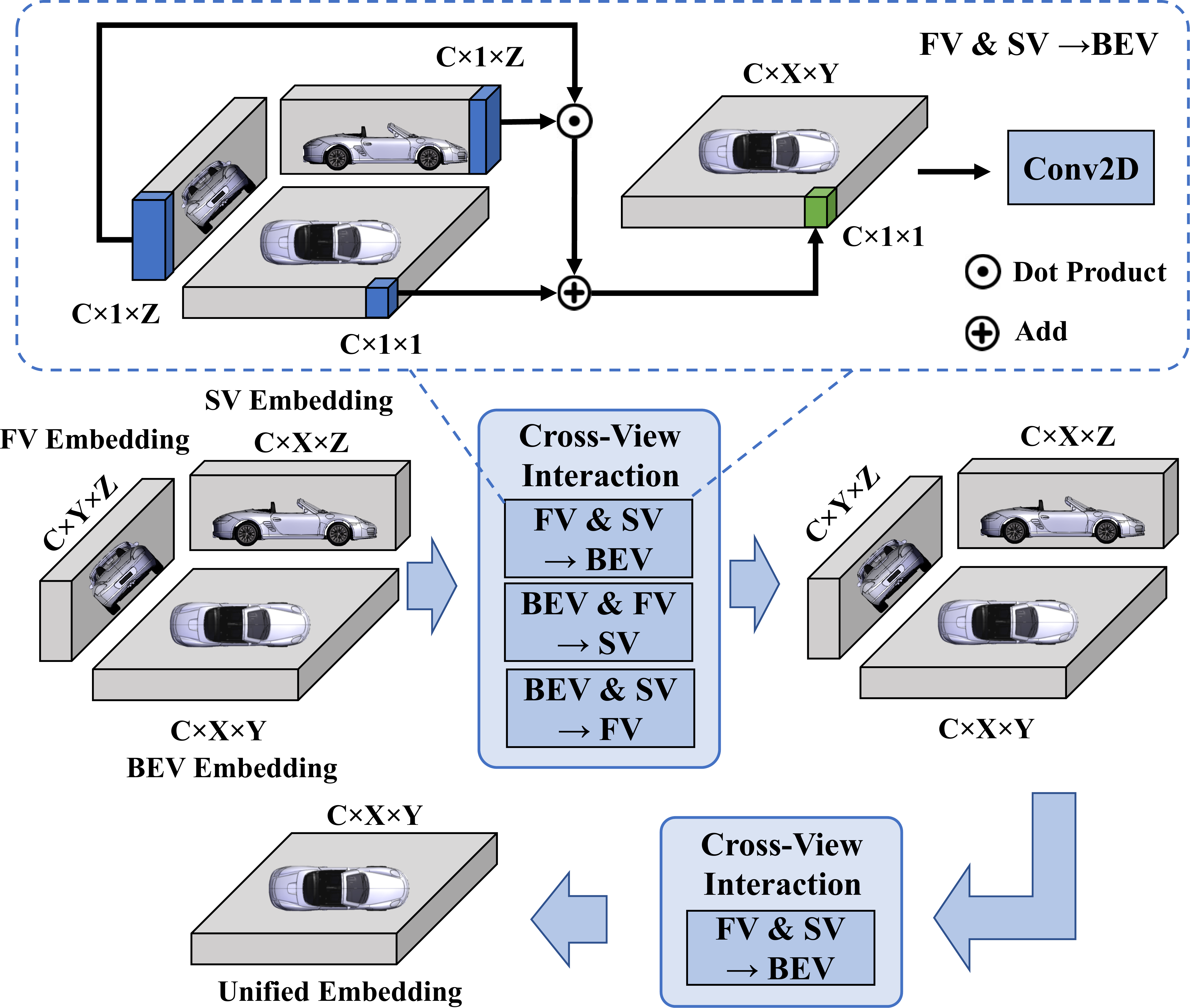}
    \caption{Illustration of Lightweight Cross-View Interaction. Matrix multiplication and 2D convolution easily fuse the embeddings with different spatial dimensions.}
    \label{fig:interact}
\end{figure}

Methods such as TPVFormer~\cite{huang2023tri} utilize multi-layer Cross-View Hybrid-Attention to aggregate TPV features, which is computationally prohibitive and contradicts our goal of maintaining the model's high efficiency. Instead, we propose a Lightweight Cross-View Interaction module, which enables interaction between embeddings of different spatial dimensions through convenient and efficient matrix multiplication. As shown in Fig.~\ref{fig:interact}, the interacted BEV Embedding $\mathbf{E}^{I}_{BEV}$ can be obtained by:
\begin{equation}   
\begin{aligned}
    \mathbf{E}^{M}_{BEV}&=(\mathbf{E}_{SV}\rightarrow \mathbb{R}^{\scriptscriptstyle C\times X\times Z})\otimes (\mathbf{E}_{FV}\rightarrow \mathbb{R}^{\scriptscriptstyle C\times Z\times Y}) \\
    \mathbf{E}^{I}_{BEV} &= {\rm Conv2D}(\mathbf{E}_{BEV} + \mathbf{E}^{M}_{BEV})
\end{aligned}
\end{equation}
The matrix multiplication $\otimes$ can also be represented as:
\begin{equation}   
    \mathbf{E}_{BEV}^{M}(x,y)=\sum_{z=1}^Z\mathbf{E}_{SV}(x,z)\cdot\mathbf{E}_{FV}(y,z)
\end{equation}
which means the explicit height information in the Z dimension of $\mathbf{E}_{FV}$ and $\mathbf{E}_{SV}$ is fused and stored in the corresponding locations of $\mathbf{E}_{BEV}$.
It is known that the elements in $\mathbf{E}_{FV}$ and $\mathbf{E}_{SV}$ have receptive fields of $X\times 3 \times 3$ and $3\times Y \times 3$ respectively. Lightweight Cross-View Interaction rapidly expands the receptive field of elements in $\mathbf{E}^{I}_{BEV}$ to $X\times Y \times Z$, covering the entire perception space.

After obtain $\mathbf{E}^{I}_{FV}$ and $\mathbf{E}^{I}_{SV}$ in the same way as $\mathbf{E}^{I}_{BEV}$,

\begin{equation}   
\begin{aligned}
    \mathbf{E}^{M}_{FV}&=(\mathbf{E}_{BEV}\rightarrow \mathbb{R}^{\scriptscriptstyle C\times Y\times X})\otimes (\mathbf{E}_{SV}\rightarrow \mathbb{R}^{\scriptscriptstyle C\times X\times Z}) \\
    \mathbf{E}^{I}_{FV} &= {\rm Conv2D}(\mathbf{E}_{FV} + \mathbf{E}^{M}_{FV})\\
    \mathbf{E}^{M}_{SV}&=(\mathbf{E}_{BEV}\rightarrow \mathbb{R}^{\scriptscriptstyle C\times X\times Y})\otimes (\mathbf{E}_{FV}\rightarrow \mathbb{R}^{\scriptscriptstyle C\times Y\times Z}) \\
    \mathbf{E}^{I}_{SV} &= {\rm Conv2D}(\mathbf{E}_{SV} + \mathbf{E}^{M}_{SV}),
\end{aligned}
\end{equation}
Lightweight Cross-View Interaction is finally used to aggregate $\mathbf{E}^{I}_{BEV}$, $\mathbf{E}^{I}_{FV}$ and $\mathbf{E}^{I}_{SV}$ into the Unified Embedding $\mathbf{E}_{U}\in \mathbb{R}^{\scriptscriptstyle C\times X\times Y}$, which can be formulated by:
\begin{equation}   
\begin{aligned}
    \mathbf{E}^{M}_{U}&=(\mathbf{E}^{I}_{SV}\rightarrow \mathbb{R}^{\scriptscriptstyle C\times X\times Z})\otimes (\mathbf{E}^{I}_{FV}\rightarrow \mathbb{R}^{\scriptscriptstyle C\times Z\times Y}) \\
    \mathbf{E}_{U} &= {\rm Conv2D}(\mathbf{E}^{I}_{BEV} + \mathbf{E}^{M}_{U})
\end{aligned}
\end{equation}
$\mathbf{E}_{U}$ encapsulates both implicit and explicit height cues of the entire driving scene, and can be seamlessly integrated with $\mathbf{F}_{BEV}$ for subsequent processing and perception. 

\subsection{Edge-aware Spatial Embedding}

Although the Spatial-to-Channel mechanism can quickly increase the receptive fields of spatial embeddings in each spatial dimension, it is not sufficiently sensitive to edge information. Therefore, we use the gradients of $\mathbf{O}_{SC}$ in each dimension to represent the edge information, which can be formulated as:
\begin{equation}
    \mathbf{O}_{SC}^{Edge}= \sqrt{\left(\frac{\partial \mathbf{O}_{SC}}{\partial x}\right)^2+\left(\frac{\partial \mathbf{O}_{SC}}{\partial y}\right)^2+\left(\frac{\partial \mathbf{O}_{SC}}{\partial z}\right)^2}
\end{equation}

By replacing $\mathbf{O}_{SC}$ with $\mathbf{O}_{SC}^{Edge}$ in Equations (3)-(9), we can obtain Unified Edge-aware Spatial Embedding $\mathbf{E}_{U}^{Edge}$ that focuses on recording edge information of the driving scene as shown in Fig.~\ref{fig:lightocc}. It not only enhances height prediction but also enables more accurate localization of vertical planes in space, thereby comprehensively improving the accuracy of occupancy prediction.

Since $\mathbf{E}_{U}$ and $\mathbf{E}_{U}^{Edge}$ share the same dimension as $\mathbf{F}_{BEV}$, they can be directly fused by
\begin{equation}
    \hat{\mathbf{F}}_{BEV}= \mathbf{F}_{BEV} + \mathbf{E}_{U} + \mathbf{E}_{U}^{Edge}.
\end{equation}
After further processing on $\hat{\mathbf{F}}_{BEV}$, the 3D semantic occupancy is predicted following the channel-to-height mechanism of FlashOcc~\cite{yu2023flashocc}.

\subsection{Geometric Supervision on Spatial Embedding}
To ensure that the obtained lightweight spatial embeddings can fully represent the spatial information in the driving scene, we utilize the ground truth occupancy to perform explicit geometric supervision on them. At the training phase, we employ an additional occupancy prediction head that takes only $\mathbf{E}_{U}$ and $\mathbf{E}_{U}^{Edge}$ as inputs to predict the geometric occupancy $\mathbf{O}_{geo}^{Pred}$. By leveraging FlashOcc's channel-to-height mechanism, the multi-layer 2D convolution prediction head maps the channel dimension $C$ of the input feature to the height dimension $Z$, which can be formulated as:
\begin{equation}
    \mathbf{O}_{geo}^{Pred}={\rm Multi\text{-}Conv2D}(\mathbf{E}_{U} + \mathbf{E}_{U}^{Edge})\rightarrow \mathbb{R}^{\scriptscriptstyle X\times Y\times Z}
\end{equation}

As shown in Fig.~\ref{fig:lightocc}, supervising $\mathbf{O}_{geo}^{Pred}$ with geometric occupancy labels generated from the ground truth occupancy can effectively enhance the ability of spatial embeddings to represent spatial information, thereby improving the final occupancy prediction accuracy. It is worth noting that the additional occupancy prediction head is not utilized during the inference phase, and thus will not affect the real-time performance of the model.

\begin{figure}[t]
    \centering

    \includegraphics[width=1\linewidth]{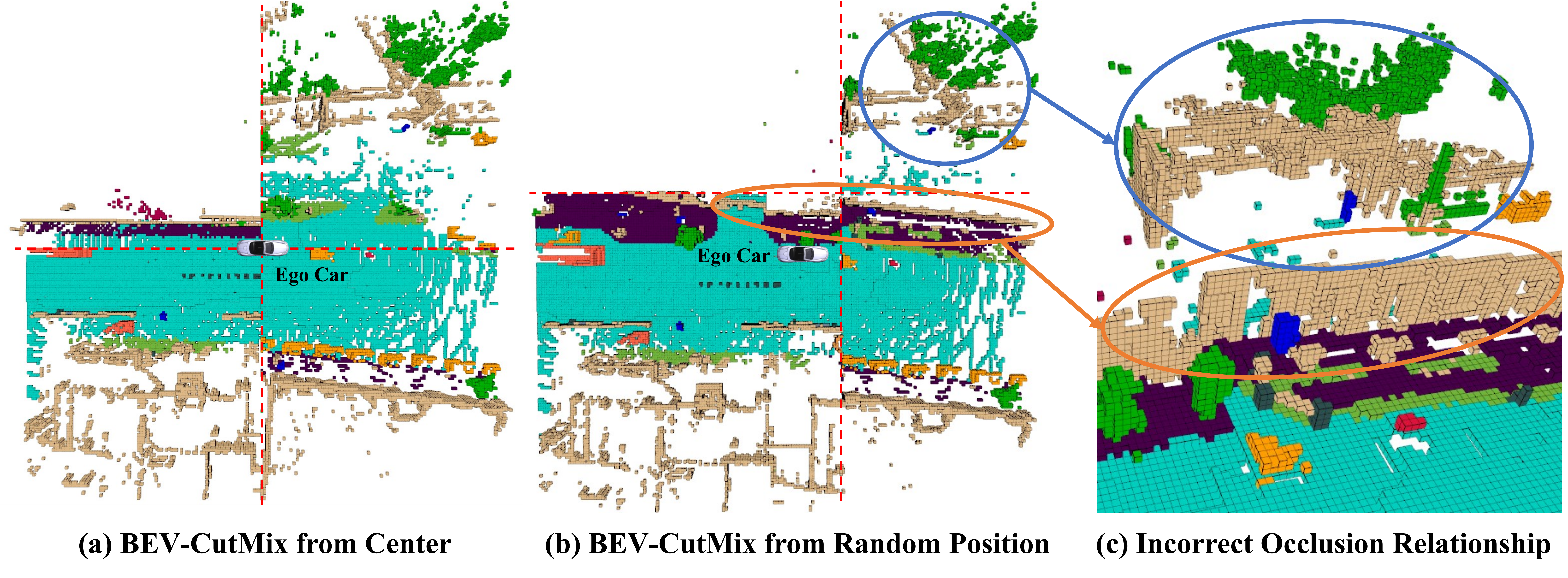} \\
    \caption{Illustration of BEV-CutMix applied on Occ3D-nuScenes dataset. We cut the occupancy into four quadrants from the center and mix these parts across scenes to construct new driving scenes while avoiding introducing incorrect occlusion relationships.}
    \label{fig:bev-cutmix}
\end{figure}

\subsection{BEV-CutMix}

Increasing the diversity of driving scenes can effectively improve the generalization performance of models~\cite{wang2024survey}. However, flexible strategies such as random scaling and random rotation are unsuitable for low-resolution and densely serialized occupancy annotations, as they often lead to spatial misalignment.
Therefore, we propose an augmentation strategy adapted to the occupancy prediction called BEV-CutMix. The occupancy labels are first cropped into several patches, after which patches from different scenes are stitched together to construct occupancy labels of the synthesized scenes. The BEV features of original scenes are also cropped and stitched in the same manner as occupancy labels to obtain the BEV features of synthesized scenes. Since occupancy labels and BEV features share identical planar coordinates, this operation is more feasible than cropping and stitching on multi-view images. 

However, there exist complex occlusion relationships among objects in the driving scene observed by the ego vehicle. Random cropping and stitching at arbitrary positions will lead to unreasonable occlusion artifacts in the synthesized scenes. As illustrated in Fig.~\ref{fig:bev-cutmix}(b) and (c), the area within the blue circle should originally be occluded by the building in the orange circle, yet it is improperly shown in the synthesized scene. 
Therefore, the cropping of occupancy labels must be centered on the ego-vehicle's position to ensure each patch contains the complete occlusion relationships. For datasets where the ego-vehicle is located at one edge of the occupancy label, such as SemanticKITTI, the label can be divided into two patches perpendicular to the edge. For datasets where the ego-vehicle is located at the center of the occupancy label, such as Occ3D-nuScenes, it can be divided into four patches along the X and Y dimensions. However, we found that dividing the label only along the X dimension yields the best results, likely because the X dimension coincides with the road direction, resulting in more realistic synthesized scenes.

\begin{table*}[h]
\centering
\footnotesize
\caption{3D occupancy prediction performance on the Occ3D-nuScenes dataset. Both the small version and the large version of LightOcc outperform the model that has similar settings. $^{\star}$ denotes the model implemented based on the official code.}
\label{tab:occ3d}
\setlength{\tabcolsep}{0.3mm}{
\begin{tabular}{l|cc|c|c|ccccccccccccccccc}
\toprule
Method
& \rotatebox{90}{History Frame}
& \rotatebox{90}{Resolution}  
& \rotatebox{90}{Backbone}      
& \rotatebox{90}{\textbf{mIoU}}  
& \rotatebox{90}{\textbf{others} }  
& \rotatebox{90}{\textbf{barrier}}  
& \rotatebox{90}{\textbf{bicycle}}  
& \rotatebox{90}{\textbf{bus}}  
& \rotatebox{90}{\textbf{car}}  
& \rotatebox{90}{\textbf{cons. veh.}}  
& \rotatebox{90}{\textbf{motorcycle}}  
& \rotatebox{90}{\textbf{pedestrian}}  
& \rotatebox{90}{\textbf{traffic cone}}  
& \rotatebox{90}{\textbf{trailer}}  
& \rotatebox{90}{\textbf{truck}}  
& \rotatebox{90}{\textbf{drive. surf.}}  
& \rotatebox{90}{\textbf{other flat}}  
& \rotatebox{90}{\textbf{sidewalk}}  
& \rotatebox{90}{\textbf{terrain}}  
& \rotatebox{90}{\textbf{manmade}}  
& \rotatebox{90}{\textbf{vegetation}}
\\ 
\midrule
OccFormer~\cite{zhang2023occformer}&  \xmark & 256$\times$704 & R50 & 20.40 & 6.6 & 32.6 & 13.1 & 20.4 & 37.1 & 5.0 & 14.0 & 21.0 & 17.0 & 9.3 & 20.6 & 40.9 & 27.0 & 27.4 & 18.7 & 18.8 & 16.9  \\ 
BEVDetOcc~\cite{huang2021bevdet}&  \xmark & 256$\times$704 & R50 & 31.64  & 6.7  & 37.0  & 8.3  & 38.7  & 44.5  & 15.2  & 13.7  & 16.4  & 15.3  & 27.1  & 31.0  & 78.7  & 36.5  & 48.3  & 51.7  & 36.8  & 32.1  \\ 
FB-Occ$^{\star}$~\cite{li2023fbocc} &  \xmark & 256$\times$704 & R50 & 35.46  & 10.8  & 39.8  & 22.0  & 43.7 & 46.3  & 22.5 & 24.6 & 25.4  & 24.7  & 28.8  & 33.8  & 77.4  & 39.0  & 46.9  & 50.2  & 35.4  & 31.6  \\ 
FlashOcc-M1~\cite{yu2023flashocc}&  \xmark & 256$\times$704 & R50 & 32.08  & 6.7  & 37.7  & 10.3  & 39.6  & 44.4  & 14.9  & 13.4  & 15.8  & 15.4  & 27.4  & 31.7  & 78.8  & 38.0  & 48.7  & 52.5  & 37.9  & 32.2  \\
OPUS-L~\cite{wang2024opus} & 8 & 256$\times$704 & R50 & 36.12 & 11.9 & 43.4 & 25.5 & 41.0 & 47.2 & \textbf{23.8} & 25.9 & 21.3 & 29.1 & 30.1 & 35.3 & 73.1 & 41.1 & 47.0 & 45.7 & 37.4 & 35.3\\
DHD-S~\cite{wu2024deep} &  \xmark & 256$\times$704 & R50 & 36.50  & 10.6  & 43.2  & 23.0  & 40.6  & 47.3  & 21.7  & 23.3  & 23.9  & 23.4  & 31.8  & 34.2  & 80.2  & 41.3  & 50.0  & 54.1  & 38.7  & 33.5 \\
\rowcolor{gray!30}
LightOcc-T &  \xmark & 256$\times$704 & R50 & 37.93  & 11.7  & 45.6  & 25.4  & 43.1  & 48.7  & 21.4  & 25.6  & 26.6  & \textbf{29.2}  & 33.2  & 35.1  & 80.0  & 41.8  & 50.4  & 53.9  & 39.4  & 34.0 \\
\rowcolor{gray!30}
LightOcc-S &  \xmark & 256$\times$704 & R50 & \textbf{39.24}  & \textbf{12.0}  & \textbf{45.9}  & \textbf{26.3}  & \textbf{47.0}  & \textbf{50.7}  & 21.7  & \textbf{27.3}  & \textbf{27.9}  & 27.3  & \textbf{34.0}  & \textbf{37.6}  & \textbf{81.2}  & \textbf{43.4}  & \textbf{52.4}  & \textbf{55.2} & \textbf{42.0}  & \textbf{35.4} \\
\midrule
TPVFormer~\cite{huang2023tri} &  \xmark & 928$\times$1600 & R101 & 27.83  & 7.2  & 38.9  & 13.7  & 40.8  & 45.9  & 17.2  & 20.0  & 18.9  & 14.3  & 26.7  & 34.2  & 55.7  & 35.5  & 37.6  & 30.7  & 19.4  & 16.8  \\ 
BEVDetOcc~\cite{huang2021bevdet}  & 1 & 512$\times$1408 & SwinB & 42.02  & 12.2 & 49.6 & 25.1 & 52.0 & 54.5 & 27.9 & 28.0 & 28.9 & 27.2 & 36.4 & 42.2 & 82.3 & 43.3 & 54.6 & 57.9 & 48.6 & 43.6 \\
FlashOcc-M3~\cite{yu2023flashocc} & 1 & 512$\times$1408 & SwinB & 43.52  & 13.4 & 51.1 & 27.7 & 51.6 & 56.2 & 27.3 & 30.0 & 29.9 & 29.8 & 37.8 & 43.5 & 83.8 & 46.6 & 56.2 & 59.6 & 50.8 & 44.7  \\ 
FastOcc~\cite{hou2024fastocc} & 16 & 640$\times$1600 & R101 & 39.21  & 12.1 & 43.5 & 28.0 & 44.8 & 52.2 & 23.0 & 29.1 & 29.7 & 27.0 & 30.8 & 38.4 & 82.0 & 41.9 & 51.9 & 53.7 & 41.0 & 35.5  \\
PanoOcc~\cite{wang2024panoocc}&  3 & 512$\times$1408 & R101 & 42.13  & 11.7 & 50.5 & 29.6  & 49.4 &  55.5  &  23.3  & 33.3 & 30.6 & 31.0 & 34.4 & 42.6 & 83.3 &  44.2 & 54.4 & 56.0 & 45.9 & 40.4  \\ 
OSP~\cite{shi2024occupancysetpoints}&  1 & 900$\times$1600 & R101 &  39.41  & 11.2 & 47.3 & 27.1 & 47.6 & 53.7 & 23.2 & 29.4 & 29.7 & 28.4 & 32.4 & 39.9 & 79.4 & 41.4 & 50.3 & 53.2 & 40.5 & 35.4  \\
COTR~\cite{ma2024cotr} &  8 & 512$\times$1408 & SwinB & 46.20  & 14.9 & 53.3 & 35.2 & 50.8 & 57.3 & \textbf{35.4} & 34.1 & 33.5 & 37.1 & 39.0 & 45.0 & 84.5 & \textbf{48.7} & \textbf{57.6} & \textbf{61.1} & 51.6 & 46.7 \\
CVT-Occ~\cite{ye2024cvt} &  8 & 928$\times$1600 & R101 & 40.34& 9.5& 49.5&23.6&49.2&55.6 &23.1 &27.9& 28.9& 29.1& 35.0& 41.0& 81.4& 40.9& 51.4&54.5&45.9 & 39.7 \\
DHD-L~\cite{wu2024deep} & 1 & 512$\times$1408 & SwinB & 45.53  & 14.1 & 53.1 & 32.4 & 52.4 & 57.4 & 30.8 & 35.2 & 33.0 & 33.4 & 37.9 & 45.3 & \textbf{84.6} & 48.0 & 57.4 & 60.3 & 52.3 & 46.2  \\ 
\rowcolor{gray!30}
LightOcc-L &  1 & 512$\times$1408 & SwinB & 46.15  & 14.5 & 52.5 & 34.5 & \textbf{53.9} & 57.5 & 31.7 & 36.1 & 34.0 & 36.3 & \textbf{40.2} & 46.2 & 84.3 & 48.3 & 57.3 & 60.3 & 51.9 & 45.4 \\ 
\rowcolor{gray!30}
LightOcc-L &  8 & 512$\times$1408 & SwinB & \textbf{47.37}  & \textbf{15.5} & \textbf{54.2} & \textbf{36.3} & 53.7 & \textbf{58.4} & 34.7 & \textbf{37.2} & \textbf{34.9} & \textbf{39.2} & 40.1 & \textbf{47.8} & 83.8 & 47.8 & 57.3 & 60.9 & \textbf{54.8} & \textbf{48.6} \\ 
\bottomrule
\end{tabular}}

\end{table*}

\begin{table*}
    \caption{Semantic scene completion results on SemanticKITTI val set. LightOcc outperforms all existing methods in mIoU. $\dag$ denotes the utilization of depth distribution predicted by the pre-trained model. $\ddag$ denotes supervision using dense depth predictions rather than sparse LiDAR depth labels.}
    \label{table:kitti_val_perf}
    \renewcommand\arraystretch{1.3}
    \setlength{\tabcolsep}{0.5mm}
    \centering
    \tiny
    \begin{tabular}{l|c c | c c c c c c c c c c c c c c c c c c c}
    \toprule
    Method
    & \textbf{IoU} & \textbf{mIoU}
    &  \rotatebox{90}{\textbf{road}}
    & \rotatebox{90}{\textbf{sidewalk}}
    & \rotatebox{90}{\textbf{parking}}
    & \rotatebox{90}{\textbf{other-grnd}}
    & \rotatebox{90}{\textbf{building}}
    & \rotatebox{90}{\textbf{car}}
    & \rotatebox{90}{\textbf{truck}}
    & \rotatebox{90}{\textbf{bicycle}}
    & \rotatebox{90}{\textbf{motorcycle}}
    & \rotatebox{90}{\textbf{other-veh.}}
    & \rotatebox{90}{\textbf{vegetation}}
    & \rotatebox{90}{\textbf{trunk}}
    & \rotatebox{90}{\textbf{terrain}}
    & \rotatebox{90}{\textbf{person}}
    & \rotatebox{90}{\textbf{bicyclist}}
    & \rotatebox{90}{\textbf{motorcyclist}}
    & \rotatebox{90}{\textbf{fence}}
    & \rotatebox{90}{\textbf{pole}}
    & \rotatebox{90}{\textbf{traf.-sign}}
    \\
    \midrule
    
    MonoScene~\cite{cao2022monoscene}  & 37.12 & 11.50 & 57.47 & 27.05 & 15.72 & 0.87 & 14.24 & 23.55 & 7.83 & 0.20 & 0.77 & 3.59 & 18.12 & 2.57 & 30.76 & 1.79 & 1.03 & 0.00 & 6.39 & 4.11 & 2.48   \\
    TPVFormer~\cite{huang2023tri}  & 35.61 &11.36 & 56.50 & 25.87 & 20.60 & 0.85 & 13.88 & 23.81 & 8.08 & 0.36 & 0.05 & 4.35 & 16.92 & 2.26 & 30.38 & 0.51 & 0.89 & 0.00 & 5.94 & 3.14 & 1.52  \\
    VoxFormer$\dag$~\cite{li2023voxformer}  & \textbf{44.02} & 12.35 & 54.76 & 26.35 & 15.50 & 0.70 & 17.65 & 25.79 & 5.63 & 0.59 & 0.51 & 3.77 & 24.39 & 5.08 & 29.96 & 1.78 & 3.32 & 0.00 & 7.64 & 7.11 & 4.18 \\
    OccFormer~\cite{zhang2023occformer} & 36.50 & 13.46 & 58.85 & 26.88 & 19.61 & 0.31 & 14.40 & 25.09 & \textbf{25.53} & 0.81 & 1.19 & 8.52 & 19.63 & 3.93 & 32.62 & 2.78 & 2.82 & 0.00 & 5.61 & 4.26 & 2.86 \\
    Symphonies$\dag$~\cite{jiang2024symphonize}& 41.92 &14.89&56.37&27.58&15.28&0.95&21.64&28.68&20.44&2.54&2.82&13.89&25.72&6.60&30.87&3.52&2.24&0.00&8.40&9.57&\textbf{5.76}\\
    OccGen~\cite{wang2024occgen}& 36.87 &13.74&61.28 &28.30 &20.42 &0.43 &14.49 &26.83 &15.49 &1.60 &2.53 &12.83 &20.04 &3.94 &32.44 &3.20 &\textbf{3.37} &0.00 &6.94 &4.11 &2.77\\
    \rowcolor{gray!30}
    LightOcc &37.25 & 14.93 & 60.08&30.58&22.80&0.11&16.04&28.10&21.88&3.88 &4.45&\textbf{17.85}&18.77&3.74&31.88&4.98&2.38&0.00&8.99&5.32&1.79\\
    \rowcolor{gray!30}
    LightOcc$\ddag$ &37.39 & 15.05 &59.72&30.34&22.50&0.16&16.99&27.97&25.25&3.55&4.27 &17.42&18.71&3.94&31.94&5.14&2.08&0.00&9.42&5.44&1.14\\
    \rowcolor{gray!30}
    LightOcc$\dag$ &43.83&\textbf{17.08}&\textbf{63.00}&\textbf{33.01}&\textbf{22.86}&\textbf{1.63}&\textbf{25.15}&\textbf{30.82}&15.10&\textbf{4.27}&\textbf{6.44}&13.98&\textbf{25.78}&\textbf{9.79}& \textbf{36.45}&\textbf{6.05}&2.50&0.00&\textbf{10.59}&\textbf{11.85}&5.22\\
    \bottomrule
    \end{tabular}
\end{table*}

\begin{table*}
    \caption{Semantic scene completion results on SemanticKITTI test set. LightOcc outperforms all existing methods in mIoU and IoU. $\dag$ denotes the utilization of depth distribution predicted by the pre-trained model.}
    \label{table:kitti_test_perf}
    \renewcommand\arraystretch{1.3}
    \setlength{\tabcolsep}{0.5mm}
    \centering
    \tiny
    \begin{tabular}{l|c c | c c c c c c c c c c c c c c c c c c c}
    \toprule
    Method
    & \textbf{IoU} & \textbf{mIoU}
    &  \rotatebox{90}{\textbf{road}}
    & \rotatebox{90}{\textbf{sidewalk}}
    & \rotatebox{90}{\textbf{parking}}
    & \rotatebox{90}{\textbf{other-grnd}}
    & \rotatebox{90}{\textbf{building}}
    & \rotatebox{90}{\textbf{car}}
    & \rotatebox{90}{\textbf{truck}}
    & \rotatebox{90}{\textbf{bicycle}}
    & \rotatebox{90}{\textbf{motorcycle}}
    & \rotatebox{90}{\textbf{other-veh.}}
    & \rotatebox{90}{\textbf{vegetation}}
    & \rotatebox{90}{\textbf{trunk}}
    & \rotatebox{90}{\textbf{terrain}}
    & \rotatebox{90}{\textbf{person}}
    & \rotatebox{90}{\textbf{bicyclist}}
    & \rotatebox{90}{\textbf{motorcyclist}}
    & \rotatebox{90}{\textbf{fence}}
    & \rotatebox{90}{\textbf{pole}}
    & \rotatebox{90}{\textbf{traf.-sign}}
    \\
    \midrule
    
    MonoScene~\cite{cao2022monoscene}  & 34.16 & 11.08 & 54.70 & 27.10 & 24.80 & 5.70  & 14.40 & 18.80 & 3.30 & 0.50 & 0.70 & 4.40 & 14.90 & 2.40  & 19.50 & 1.00 & 1.40 & 0.40 & 11.10 & 3.30 & 2.10   \\
    TPVFormer~\cite{huang2023tri}  & 34.25 & 11.26 & 55.10 & 27.20 & 27.40 & 6.50  & 14.80 & 19.20 & 3.70 & 1.00 & 0.50 & 2.30 & 13.90 & 2.60  & 20.40 & 1.10 & 2.40 & 0.30 & 11.00 & 2.90 & 1.50  \\
    VoxFormer$\dag$~\cite{li2023voxformer}  & 42.95 & 12.20 & 53.90 & 25.30 & 21.10 & 5.60  & 19.80 & 20.80 & 3.50 & 1.00 & 0.70 & 3.70 & 22.40 & 7.50  & 21.30 & 1.40 & 2.60 & 0.20 & 11.10 & 5.10 & 4.90 \\
    OccFormer~\cite{zhang2023occformer} & 34.53 & 12.32 & 55.90 & 30.30 & 31.50 & 6.50  & 15.70 & 21.60 & 1.20 & 1.50 & 1.70 & 3.20 & 16.80 & 3.90  & 21.30 & 2.20 & 1.10 & 0.20 & 11.90 & 3.80 & 3.70  \\
    Symphonies$\dag$~\cite{jiang2024symphonize}& 42.19 & 15.04 & 58.40 & 29.30 & 26.90 & 11.70 & 24.70 & 23.60 & 3.20 & 3.60 & 2.60 & 5.60 & 24.20 & 10.00 & 23.10 & 3.20 & 1.90 & \textbf{2.00} & 16.10 & 7.70 & 8.00\\
    CGFormer$\dag$~\cite{cgformer}&44.41&16.63&64.30&34.20&34.10&12.10&25.80 &26.10&4.30&3.70&1.30&2.70&24.50&11.20& 29.30&1.70&3.60&0.40&18.70&8.70&\textbf{9.30}\\
    \rowcolor{gray!30}
    LightOcc$\dag$ &\textbf{44.63}&\textbf{17.92}&\textbf{64.60}&\textbf{34.30}&\textbf{36.00}&\textbf{15.40}&\textbf{28.00}&\textbf{26.60}&\textbf{5.00}&\textbf{5.50}&\textbf{5.00}&\textbf{8.20}&\textbf{24.90}&\textbf{11.90}&\textbf{28.80}&\textbf{4.20}&\textbf{5.20}&0.50&\textbf{19.50}&\textbf{10.00}&6.90\\
    \bottomrule
    \end{tabular}
\end{table*}

\begin{table*} %
    \caption{3D semantic occupancy prediction results on SurroundOcc Benchmark. Our method achieves state-of-the-art performance in both IoU and mIoU.}
    \label{tab:surroundocc}
    \setlength{\tabcolsep}{0.7mm}  
    \tiny
    \renewcommand\arraystretch{1.3}
    \centering
    \begin{tabular}{l|c c | c c c c c c c c c c c c c c c c}
        \toprule
        Method
        &  \textbf{IoU} & \textbf{mIoU}
        & \rotatebox{90}{\textbf{barrier}}
        & \rotatebox{90}{\textbf{bicycle}}
        & \rotatebox{90}{\textbf{bus}}
        & \rotatebox{90}{\textbf{car}}
        & \rotatebox{90}{\textbf{const. veh.}}
        & \rotatebox{90}{\textbf{motorcycle}}
        & \rotatebox{90}{\textbf{pedestrian}}
        & \rotatebox{90}{\textbf{traffic cone}}
        & \rotatebox{90}{\textbf{trailer}}
        & \rotatebox{90}{\textbf{truck}}
        & \rotatebox{90}{\textbf{drive. suf.}}
        & \rotatebox{90}{\textbf{other flat}}
        & \rotatebox{90}{\textbf{sidewalk}}
        & \rotatebox{90}{\textbf{terrain}}
        & \rotatebox{90}{\textbf{manmade}}
        & \rotatebox{90}{\textbf{vegetation}}
        \\
        \midrule
        MonoScene~\cite{cao2022monoscene} & 23.96 & 7.31 & 4.03 &	0.35& 8.00& 8.04&	2.90& 0.28& 1.16&	0.67&	4.01& 4.35&	27.72&	5.20& 15.13&	11.29&	9.03&	14.86 \\
        Atlas~\cite{murez2020atlas} & 28.66 & 15.00 & 10.64&	5.68&	19.66& 24.94& 8.90&	8.84&	6.47& 3.28&	10.42&	16.21&	34.86&	15.46&	21.89&	20.95&	11.21&	20.54 \\ 
        BEVFormer~\cite{li2022bevformer} & 30.50 & 16.75 & 14.22 &	6.58 & 23.46 & 28.28& 8.66 &10.77& 6.64& 4.05& 11.20&	17.78 & 37.28 & 18.00 & 22.88 & 22.17 & 13.80 &	22.21\\
        TPVFormer~\cite{huang2023tri}  & 30.86 & 17.10 & 15.96&	 5.31& 23.86	& 27.32 & 9.79 & 8.74 & 7.09 & 5.20& 10.97 & 19.22 & 38.87 & 21.25 & 24.26 & 23.15 & 11.73 & 20.81\\
        OccFormer~\cite{zhang2023occformer} & 31.39 & 19.03 & 18.65 & 10.41 & 23.92 & 30.29 & 10.31 & 14.19 & 13.59 & 10.13 & 12.49 & 20.77 & 38.78 & 19.79 & 24.19 & 22.21 & 13.48 & 21.35\\
        SurroundOcc~\cite{wei2023surroundocc} & 31.49 & 20.30  & 20.59 & 11.68 & 28.06 & 30.86 & 10.70 & 15.14 & 14.09 & 12.06 & 14.38 & 22.26 & 37.29 & 23.70 & 24.49 & 22.77 & 14.89 & 21.86  \\ 
        LowRankOcc~\cite{zhao2024lowrankocc} & 32.78 &21.51 &22.49 &12.45 &30.32 &\textbf{33.63} &10.35 &14.31 &13.67 &12.40 &15.09 &\textbf{25.99} &39.52& 23.21 &26.67 &25.19 &16.23 &22.66 \\
        GaussianFormer~\cite{huang2024gaussianformer} & 29.83 & 19.10 & 19.52 & 11.26 & 26.11 & 29.78 & 10.47 & 13.83 & 12.58 & 8.67 & 12.74 & 21.57 & 39.63 & 23.28 & 24.46 & 22.99 & 9.59 & 19.12 \\
        GaussianFormer-2~\cite{huang2025gaussianformer} &31.74 &20.82 &21.39 &13.44 &28.49& 30.82 &10.92 &15.84 &13.55 &10.53 &14.04 &22.92 &40.61 &24.36 &26.08 &24.27 &13.83 &21.98 \\
        GaussianWorld~\cite{zuo2025gaussianworld} & 33.40 &22.13 &21.38 &14.12 &27.71 &31.84 &13.66 &17.43 &13.66 &11.46 &15.09 & 23.94 &\textbf{42.98} &24.86 &28.84 &26.74 &15.69 &\textbf{24.74}\\
        \rowcolor{gray!30}
        LightOcc &\textbf{34.01} & \textbf{23.68}&\textbf{24.65}&\textbf{15.37}&\textbf{31.02}&33.48&\textbf{14.27} &\textbf{19.42}&\textbf{17.17}&\textbf{15.21}&\textbf{16.81}&25.63&42.47&\textbf{26.31}&\textbf{29.15}&\textbf{27.35}&\textbf{15.85}& \textbf{24.74}\\
        \bottomrule
    \end{tabular}
\end{table*}

\begin{table*} %
    \caption{3D semantic occupancy prediction results on OpenOccupancy Benchmark. Our method achieves state-of-the-art performance in both IoU and mIoU.}
    \label{tab:openocc}
    \setlength{\tabcolsep}{0.6mm}  
    \footnotesize
    \renewcommand\arraystretch{1.3}
    \centering
    \begin{tabular}{l|c|c c | c c c c c c c c c c c c c c c c}
        \toprule
        Method & Modality
        &  \textbf{IoU} & \textbf{mIoU}
        & \rotatebox{90}{\textbf{barrier}}
        & \rotatebox{90}{\textbf{bicycle}}
        & \rotatebox{90}{\textbf{bus}}
        & \rotatebox{90}{\textbf{car}}
        & \rotatebox{90}{\textbf{const. veh.}}
        & \rotatebox{90}{\textbf{motorcycle}}
        & \rotatebox{90}{\textbf{pedestrian}}
        & \rotatebox{90}{\textbf{traffic cone}}
        & \rotatebox{90}{\textbf{trailer}}
        & \rotatebox{90}{\textbf{truck}}
        & \rotatebox{90}{\textbf{drive. suf.}}
        & \rotatebox{90}{\textbf{other flat}}
        & \rotatebox{90}{\textbf{sidewalk}}
        & \rotatebox{90}{\textbf{terrain}}
        & \rotatebox{90}{\textbf{manmade}}
        & \rotatebox{90}{\textbf{vegetation}}
        \\
        \midrule
        MonoScene~\cite{cao2022monoscene}& Camera & 18.4 & 6.9 & 7.1  & 3.9  &  9.3 &  7.2 & 5.6  & 3.0  &  5.9& 4.4& 4.9 & 4.2 & 14.9 & 6.3  & 7.9 & 7.4  & 10.0 & 7.6  \\
        TPVFormer~\cite{huang2023tri} &Camera & 15.3 &  7.8 & 9.3  & 4.1  &  11.3 &  10.1 & 5.2  & 4.3  & 5.9 & 5.3&  6.8& 6.5 & 13.6 & 9.0  & 8.3 & 8.0  & 9.2 & 8.2 \\
        OpenOccupancy \cite{wang2023openoccupancy} &Camera&19.3  & 10.3  &  9.9 & 6.8  & 11.2  & 11.5  & 6.3  & 8.4  & 8.6 & 4.3 & 4.2 & 9.9 & 22.0  & 15.8 & 14.1  & 13.5  & 7.3&10.2 \\
        CONet \cite{wang2023openoccupancy} & Camera &20.1  & 12.8&13.2  & 8.1 & \textbf{15.4} &  17.2 & 6.3  & 11.2  & 10.0  &  8.3 & 4.7 & 12.1 & 31.4 & 18.8 & 18.7  & 16.3 & 4.8  &8.2  \\ 
        OccGen~\cite{wang2024occgen}&Camera& 23.4 & 14.5 & 15.5&  9.1 & 15.3  & \textbf{19.2} & 7.3 & 11.3& 11.8 & \textbf{8.9}& 5.9& 13.7&\textbf{34.8}& 22.0& 21.8& 19.5 & 6.0& 9.9\\
        OccMamba~\cite{li2025occmamba}&Camera&21.7 &13.2 &13.2 &7.4 &15.3 &17.6 &6.1 &10.3 &10.3 &6.5 &6.2 &13.3 &32.9 &20.5 &19.8 &17.4 &4.8 &8.7 \\
        \rowcolor{gray!30}
        LightOcc & Camera &\textbf{25.4} & \textbf{15.6}&\textbf{16.2}&\textbf{9.7}&15.1&18.5 &\textbf{9.2}&\textbf{12.3}&\textbf{12.1}&8.3&\textbf{7.8}&\textbf{14.2}&33.1&\textbf{22.5}&\textbf{22.7}&\textbf{20.9}&\textbf{11.4}& \textbf{15.9}\\
        \bottomrule
    \end{tabular}
\end{table*}

\section{Experiments}

\subsection{Benchmark and Evaluation Metric}
Occ3D-nuScenes~\cite{tian2024occ3d} is built upon nuScenes dataset~\cite{caesar2020nuscenes}, the commonly used large-scale autonomous driving dataset, which contains 700 scenes for training and 150 scenes for validation. The occupancy annotations cover a spatial range from -40m to 40m along the X and Y dimensions, and -1m to 5.4m along the Z dimension. Each voxel in the occupancy annotations is a cube with a length of 0.4m. For evaluating the occupancy prediction performance, the mean intersection-over-union (mIoU) of all semantic categories is employed. 

SemanticKITTI~\cite{behley2019semantickitti} provides semantic occupancy annotations for the data of KITTI Odometry Benchmark~\cite{geiger2012we}. The ground-truth has a resolution of 256$\times$256$\times$32 voxel grids, with each voxel spanning 0.2m. In addition to mIoU, non-semantic IoU is also utilized to evaluate model performance.

SurroundOcc~\cite{wei2023surroundocc} Benchmark is also built upon nuScenes. Unlike Occ3D-nuScenes~\cite{tian2024occ3d}, SurroundOcc adopts the LiDAR coordinate system and covers a spatial range from -50m to 50m along the X and Y dimensions, and -5m to 3m along the Z dimension. Each annotated voxel has a length of 0.5m. Both IoU and mIoU are used to evaluate model performance, with the ``other'' category excluded during evaluation.

OpenOccupancy~\cite{wang2023openoccupancy} Benchmark is also a commonly used occupancy prediction baseline, which is annotated based on the nuScenes dataset. It adopts a perceptual range similar to that of SurroundOcc, but has more fine-grained annotations. The occupancy annotations are represented in a 512$\times$512$\times$40 voxel grid, with each voxel sized at 0.2 meters. The adopted metrics are the same as SurroundOcc benchmark. 

\subsection{Implementation Details}
We adopt the FlashOcc~\cite{yu2023flashocc} optimized by commonly used modifications as the baseline and apply our Lightweight Spatial Embedding to supplement BEV features with height information. When evaluating on Occ3D-nuScenes benchmark, we implement three versions of LightOcc, namely LightOcc-T, LightOcc-S and LightOcc-L, to achieve the best trade-off between accuracy and efficiency, respectively. LightOcc-T aligns with FlashOcc-M1, employing the ResNet-50~\cite{he2016deep} as the image backbone and processing images at a resolution of $256\times 704$. LightOcc-S adds a depth prediction module on the basis of LightOcc-T, which is supervised by depth labels generated by LiDAR.  The model size of LightOcc-L is comparable to FlashOcc-M3, which utilizes Swin-Transformer-Base~\cite{liu2021swin} as the image backbone and inputs the images in the size of $512\times 1408$. The Multi-view Stereo module~\cite{li2023bevstereo} used by FlashOcc-M3 is also employed. For the ablation study, we conduct experiments on LightOcc-T or LightOcc-S and train for 24 epochs without BEV-CutMix for quick verification.

When evaluating on other benchmarks, we adopt the LightOcc-L configuration for experiments to compare with more state-of-the-art occupancy prediction methods, thereby comprehensively demonstrating the performance of LightOcc. For OpenOccupancy benchmark with higher occupancy resolution, LightOcc directly predicts the occupancy in $\frac{1}{2}\times\frac{1}{2}\times\frac{1}{2}$ resolution and upsample it to the original resolution for metrics calculation.

\subsection{Main Results}
The experiment results on Occ3D-nuScenes benchmark are shown in Tab.~\ref{tab:occ3d}. It can be found that all of LightOcc-T, LightOcc-S and LightOcc-L outperform other approaches with comparable model configurations. LightOcc-T improves the performance of the BEV-based FlashOcc-M1 model, by 5.85\% mIoU. LightOcc-S outperforms state-of-the-art, DHD-S~\cite{wu2024deep}, by 2.74\%. LightOcc-L adopts a heavier configuration to pursue higher accuracy. When utilizing 1 history frame, Light-L improves the performance of FlashOcc-M3 model by 2.63\% mIoU. When utilizing 8 history frames, LightOcc-L outperforms COTR~\cite{ma2024cotr} by 1.17\% mIoU. These persuasive experimental results highlight the effectiveness of Lightweight Spatial Embedding in supplementing the height information to the BEV features. 

The experiment results on SemanticKITTI val dataset are shown in Tab.~\ref{table:kitti_val_perf}. When adopting depth distributions predicted by the model itself, LightOcc outperforms OccGen~\cite{wang2024occgen} by 1.19\% mIoU. Following methods such as VoxFormer~\cite{li2023voxformer} and Symphonies~\cite{jiang2024symphonize}, when adopting depth distributions estimated from binocular disparity by MobileStereoNet~\cite{shamsafar2022mobilestereonet}, LightOcc achieves obvious performance improvement and outperforms Symphonies by 2.19\% mIoU and 1.91\% IoU. Besides, supervising LightOcc's depth module with MobileStereoNet's depth predictions can also increase the accuracy compared to the model trained with sparse LiDAR depth labels. 
We also conduct experiments on the SemanticKITTI test dataset and show the results in Tab.~\ref{table:kitti_test_perf}. LightOcc also achieves the best performance and outperforms Symphonies by 2.88\% mIoU/2.44\%IoU and CGFormer~\cite{cgformer} by 1.29\%mIoU/0.22\%IoU.

The experiment results on SurroundOcc benchmark are shown in Tab.~\ref{tab:surroundocc}. Our proposed LightOcc achieves the state-of-the-art performance in both IoU and mIoU. Specifically, LightOcc outperforms LowRankOcc~\cite{zhao2024lowrankocc} by 1.23\% IoU / 2.17\% mIoU and outperforms GaussianWorld~\cite{zuo2025gaussianworld} by 0.61\% IoU / 1.55\% mIoU. 

For OpenOccupancy benchmark, some methods utilize multi-modal inputs, and we only conduct a fair comparison with their camera-only models and present the results in Tab.~\ref{tab:openocc}. It can be found that LightOcc achieves the best IoU and mIoU metrics. Specifically, LightOcc outperforms OccMamba~\cite{li2025occmamba} by 3.7\% IoU / 2.4\% mIoU and outperforms OccGen~\cite{wang2024occgen} by 2.0\% IoU / 1.1\% mIoU. These experimental results further highlight the performance advantage of LightOcc.

\begin{table*}[t]
    \caption{Efficiency comparison with other methods. ``FlashOcc-opt'' denotes the optimized FlashOcc model, with its details presented in Tab.~\ref{tab:opt_baseline}. ``Voxel Baseline'' denotes the model where the LightOcc modules proposed in LightOcc are replaced with standard voxel feature processing modules. $^{\star}$ denotes the model implemented based on the official code. FPS are evaluated on a single RTX3090 GPU.}
    \label{tab:efficiency}
    \centering
    \footnotesize
    \setlength\tabcolsep{1.5mm}
    \begin{tabular}{l|c|cc|c|c|cccc}
        \hline
        Method & Representation & Image Size & Backbone & History Frame & mIoU & Params & FLOPs &Memory & FPS  \\
        \hline
BEVDetOcc~\cite{huang2021bevdet} & Voxel & 256$\times$704 & R50 & \xmark & 31.64 & 29.02M & 241.76G & 1.43GB& 15.60 \\
FB-Occ$^{\star}$~\cite{li2023fbocc} & BEV+Voxel & 256$\times$704 & R50 & \xmark & 35.46 & 68.36M & 510.55G & 1.50GB& 5.43 \\
FlashOcc-M1~\cite{yu2023flashocc} & BEV & 256$\times$704 & R50 & \xmark & 32.08 & 44.74M & 248.57G & 0.50GB& 28.51 \\
FlashOcc-opt & BEV & 256$\times$704 & R50 & \xmark & 35.00 & 48.87M & 227.92G & 0.47GB & 29.84 \\
DHD-S~\cite{wu2024deep} & Multi-BEV & 256$\times$704 & R50 & \xmark & 36.50 & 146.69M & 445.06G & 0.60GB& 9.95 \\
COTR$^{\star}$~\cite{ma2024cotr}& Compact Voxel & 256$\times$704 & R50 & \xmark & 36.90 & 33.57M & 259.69G & 2.85GB& 7.99\\
CVT-Occ$^{\star}$~\cite{ye2024cvt}& 3D Cost Volume & 480$\times$800 & R50 & 1 & 32.66 &43.71M& 357.82G& 0.85GB&9.52 \\
OPUS$^{\star}$~\cite{wang2024opus} & Sparse Query &  256$\times$704 & R50 & \xmark & 27.31 & 71.96M& 176.29G& 0.58GB & 14.60 \\
\hline
Voxel Baseline-T & Voxel & 256$\times$704 & R50 & \xmark & 36.55 & 63.81M & 573.38G & 1.22GB& 10.98 \\
Voxel Baseline-S & Voxel & 256$\times$704 & R50 & \xmark & 38.74 &  69.19M & 594.28G & 1.26GB& 10.51\\
\hline
LightOcc-T & BEV & 256$\times$704 & R50 & \xmark & 37.93 & 49.35M & 235.52G & 0.52GB& 28.38 \\
LightOcc-S & BEV & 256$\times$704 & R50 & \xmark & 39.24 & 55.20M & 264.07G & 0.52GB& 24.76\\
\hline
    \end{tabular}
\end{table*}

\begin{table}
    \caption{Details about the optimized FlashOcc. ``RC-Sampling'' denotes using RC-Sampling~\cite{zhang2024geobev} for view transformation. ``BEV FPN'' denotes optimizing the structure of the BEV FPN. ``Auxiliary Loss'' denotes the additional losses used by FB-Occ~\cite{li2023fbocc} besides Cross-Entropy Loss.}
    \label{tab:opt_baseline}
    \centering
    \footnotesize
    \setlength{\tabcolsep}{1mm}{
    \begin{tabular}{ccc|ccc}
        \hline
        RC-Sampling & BEV FPN & Auxiliary Loss & mIoU$\uparrow$ & Latency$\downarrow$\\ 
        \hline
& & & 32.08 & 35.08 ms\\
        \cmark & & & 32.70 & 34.72 ms\\
        \cmark & \cmark & & 32.83 &33.51 ms \\
        \cmark & \cmark & \cmark & 35.00 &33.51 ms \\
        \hline
    \end{tabular}}
\end{table}

\subsection{Efficiency Performance}
Tab.~\ref{tab:efficiency} presents the comparison of efficiency among several models with small scale, which reflects the practical deployment capabilities of these methods. We optimize FlashOcc to demonstrate the comprehensive advantages of BEV-based methods in terms of accuracy and efficiency. The optimizations shown in Tab.~\ref{tab:opt_baseline} include replacing BEVPoolv2~\cite{huang2022bevpoolv2} with RC-Sampling~\cite{zhang2024geobev} to obtain BEV feature with higher geometry quality, modifying BEV FPN module to reduce computational cost. More importantly, Scene-Class Affinity Loss~\cite{cao2022monoscene} and  Lov{\'a}sz-Softmax Loss~\cite{berman2018lovasz} utilized by FB-Occ~\cite{li2023fbocc} are incorporated with Cross-Entropy Loss. FlashOcc-opt only lags behind FB-Occ by 0.46\% mIoU, while achieving its $5.5\times$ FPS.

It can be found in Tab.~\ref{tab:efficiency} that LightOcc-T maintains the high efficiency of FlashOcc-opt while improving mIoU by 2.93\%, demonstrating the effectiveness and efficiency Lightweight Spatial Embeddings. When comparing with the voxel-based methods, LightOcc-S outperforms FB-Occ~\cite{li2023fbocc} by 3.78\% mIoU, while achieving its $4.56\times$ FPS. Compared to DHD-S~\cite{wu2024deep}, which uses multiple BEV features to capture scene information at different heights, LightOcc also holds an advantage of 2.74\% mIoU, while achieving its $2.5\times$ FPS. We also compare with the latest efficiency-oriented methods, such as CVT-Occ~\cite{ye2024cvt}, COTR~\cite{ma2024cotr}, OPUS~\cite{wang2024opus}. LightOcc not only achieves higher accuracy but also has a clear advantage in efficiency. 
By replacing the lightweight modules of LightOcc with standard 3D modules, the Voxel Baseline models are obtained as references. Their accuracy and efficiency are both decreased compared to LightOcc, demonstrating the effectiveness of our proposed lightweight modules.

\subsection{Ablation Study}

\subsubsection{Effectiveness of Components}
\begin{table}[t]
    \caption{Ablation study of components used in LightOcc. ``S2C'' denotes Spatial-to-Channel Mechanism. ``LCVI'' denotes  Lightweight Cross-View Interaction. ``Edge'' denotes Edge-Aware Spatial Embeddings. ``GS'' denotes the Geometry Supervision on Spatial Embeddings. ``CutMix'' denotes training more epochs with BEV-CutMix.}
    \label{tab:components}
    \centering
    \footnotesize
    \setlength{\tabcolsep}{1.2mm}{
    \begin{tabular}{l|ccccc|cc}
        \hline
        Model & S2C & LCVI & Edge & GS & CutMix & mIoU \\ 
        \hline
        \multirow{2}{*}{FlashOcc-opt} & & & & & & 35.00 \\
         & & & & & \cmark & 35.89 \\
        \hline
        \multirow{3}{*}{LightOcc-T}& \cmark & & & & & 36.30 \\
        & \cmark & \cmark & & & & 36.86 \\
        & \cmark & \cmark & & & \cmark & 37.93 \\
        \hline
        \multirow{4}{*}{LightOcc-S} & \cmark & \cmark & & & & 38.17 \\
        & \cmark & \cmark & \cmark & & & 38.46 \\
        & \cmark & \cmark & \cmark & \cmark & & 38.70 \\
        & \cmark & \cmark & \cmark & \cmark & \cmark & 39.24 \\
        \hline
    \end{tabular}}
\end{table}

We evaluate the effectiveness of each proposed module through experiments on the Occ3D-nuScenes dataset, with results presented in the Tab.~\ref{tab:components}. Firstly, adopting the Spatial-to-Channel mechanism to acquire BEV embeddings effectively improves model accuracy by 1.30\% mIoU. Subsequently, the Lightweight Cross-View Interaction module facilitates interaction and fusion of spatial embeddings from different views, further boosting performance by 0.56\%mIoU. After constructing the LightOcc-S model with an additional depth branch, the extraction of Edge-Aware Spatial Embedding and the Geometric Supervision applied on embeddings separately contribute to accuracy by 0.29\% and 0.24\% mIoU. Besides, BEV-CutMix, as a highly generalizable data augmentation strategy, achieves consistent performance improvements across various models.

\subsubsection{Lightweight Spatial Embedding}

We conduct the ablation study of the Lightweight Spatial Embedding, including Spatial-to-Channel mechanism and Lightweight Cross-view Interaction module. From the result shown in Tab.~\ref{tab:embedding}, it can be found that only utilizing $\mathbf{E}_{BEV}$ without Cross-view Interaction can improve mIoU by 1.30\%. After applying Lightweight Cross-View Interaction, the implicit and explicit height information carried by $\mathbf{E}_{BEV}$, $\mathbf{E}_{FV}$ and $\mathbf{E}_{SV}$ are fused in $\bold{E}_{U}$, which can further improve the accuracy performance. We adopt 2 interaction layers as a balance between accuracy and efficiency, which brings another 0.56\% mIoU improvement while only increasing latency by 0.42 ms. It demonstrates the high efficiency of the Lightweight Cross-View Interaction module. If each interaction layer contains two convolutions, the loss of efficiency does not result in a significant improvement in accuracy, which indicates that the sparse height information can be effectively processed using a single layer of convolution. 

We also attempt to use the Cross-View Hybrid-Attention module proposed by TPVFormer~\cite{huang2023tri} to achieve interaction between $\mathbf{E}_{BEV}$, $\mathbf{E}_{FV}$ and $\mathbf{E}_{SV}$. The results show that although it can also improve the final accuracy, it increases the latency by over 10 ms. This indicates that Lightweight Cross-View Interaction module shows better practicality when processing spatial embeddings that contain only spatial information.

Other factors that affect the quality of Lightweight Spatial Embedding are shown in Tab.~\ref{tab:embedding_supp}. It can be found that the activation function utilized to obtain multi-view depth distributions can largely affect the final accuracy. Specifically, Softmax tends to centralize the weight on a specific depth value, which does not match the representation required by occupancy prediction. On the contrary, Sigmoid maps depth weight at each depth value into [0,1], which can be regarded as the probability of being occupied. Single-channel occupancy sampled from the sigmoid-activated depth distribution can better represent the spatial information of the scene. Besides, the multiplication between embeddings will vanish their common dimension and lead to value accumulation. Dividing the multipled embeddings by the vanished spatial dimension can balance their values with the original embeddings and bring 0.15\% mIoU improvement.

\begin{table}[t]
    \caption{Ablation study of the module structure to obtain Lightweight Spatial Embedding. ``LCVI'' denotes our proposed Lightweight Cross-View Interaction. ``CVHA'' denotes the Cross-View Hybrid-Attention utilized in TPVFormer. ``Layers'' is the number of LCVI layers. ``Convs'' is the number of convolutions used each time for embedding extraction and interaction.}
    \label{tab:embedding}
    \centering
    \footnotesize
    \setlength{\tabcolsep}{3mm}{
    \begin{tabular}{c|cc|cc}
        \hline
        Interaction & Layers & Convs & mIoU & Latency  \\ 
        \hline
        \multirow{2}{*}{None} & - & - & 35.00 &  33.51 ms \\
& - & 2  & 36.30 &  34.81 ms \\
\hline
\multirow{4}{*}{LCVI}& 1 & 1  & 36.60 &  35.06 ms \\
& 2 & 1 & 36.86 & 35.23 ms \\
& 2 & 2  & 36.92 & 36.40 ms\\
&  3  & 1 & 36.82 &  35.98 ms \\
\hline
CVHA  & - & - & 36.60 & 45.63 ms\\
        \hline
    \end{tabular}}
\end{table}

\begin{table}[t]
    \caption{Ablation study of Lightweight Spatial Embedding. ``Depth Act'' denotes the activation function used by depth distribution. ``Mean'' denotes to average the multiplied embeddings by the vanished dimensions.}
    \label{tab:embedding_supp}
    \centering
    \setlength{\tabcolsep}{1.8mm}
    \begin{tabular}{l|c|c|cc}
        \hline
        Model & Depth Act & Mean & mIoU$\uparrow$ & Latency$\downarrow$\\ 
        \hline
        \multirow{3}{*}{LightOcc-T}&Softmax & \xmark & 36.11 & 35.23 ms\\
        &Sigmoid & \xmark & 36.71 & 35.23 ms\\
        &Sigmoid & \cmark & 36.86 & 35.23 ms\\
        \hline
    \end{tabular}
\end{table}

\begin{table}
    \caption{Ablation study of Edge-aware Spatial Embedding and Geometric Supervision. ``Edge'' denotes the way of calculating the edge of Single-Channel Occupancy. ``Supervision'' denotes the Geometric Supervision on Spatial Embedding.}
    \label{tab:edge}
    \centering
    \footnotesize
    \setlength{\tabcolsep}{0.5mm}{
    \begin{tabular}{l|c|c|ccc}
        \hline
        Model & Edge & Supervision & mIoU$\uparrow$ & IoU$\uparrow$  & Latency$\downarrow$\\ 
        \hline
        \multirow{5}{*}{LightOcc-S}& - & \xmark & 38.17 & 67.85 &38.46 ms\\
        & - & \cmark & 38.54 & 68.35& 38.46 ms\\
& Gradient & \xmark & 38.46 & 68.01 & 40.38 ms\\
& Gradient & \cmark & 38.70 &68.61 & 40.38 ms\\
& Sobel & \xmark & 38.56 & 68.56& 42.57 ms\\
        \hline
    \end{tabular}}
\end{table}

\begin{table}[t]
    \caption{Additional experiment about BEV-CutMix. All model are trained for 48 epochs.}
    \label{tab:supp_bev_cutmix}
    \centering
    \footnotesize
    \setlength{\tabcolsep}{3mm}{
    \begin{tabular}{l|c}
        \hline
        Strategy & mIoU$\uparrow$ \\ 
        \hline
         w/o BEV-CutMix  & 37.27 \\
Random Cutting and Mixing & 37.59 \\
Cutting from Center along X and Y axis & 37.75 \\
Cutting from Center along Y axis & 37.93 \\
        \hline
    \end{tabular}}
\end{table}

\begin{figure*}[t]
    \centering
    \includegraphics[width=1\linewidth]{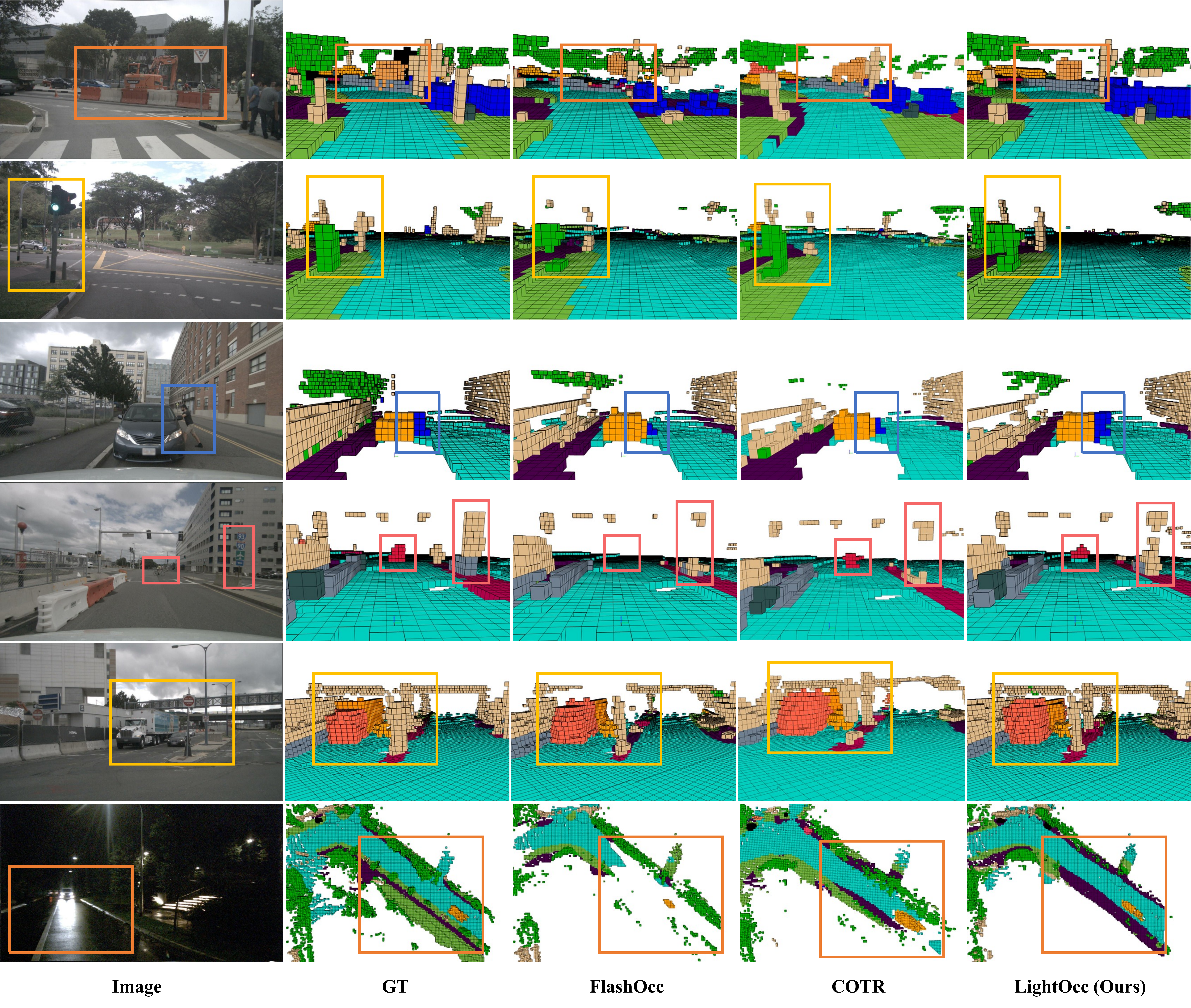}
    \caption{Visualization results of LightOcc, FlashOcc~\cite{yu2023flashocc} and COTR~\cite{ma2024cotr} on Occ3D-nuScenes benchmark. The rectangles illustrate that Lightweight Spatial Embeddings can effectively supplement the height clues and improve the accuracy of height predictions.}
    \label{fig:visualization}
\end{figure*}

\begin{figure}[t]
    \centering
    \includegraphics[width=1\linewidth]{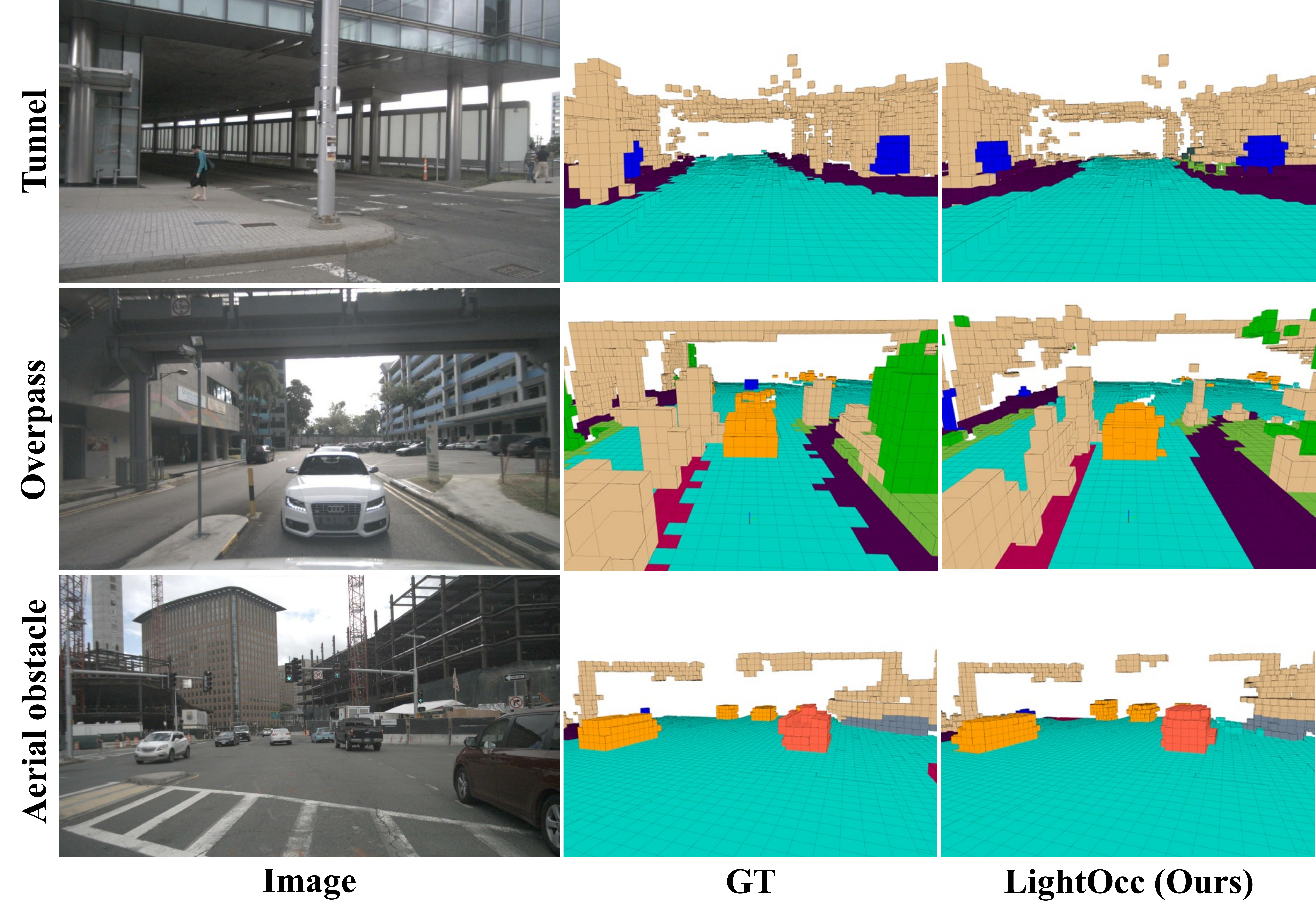}
    \caption{Visualization results of LightOcc on scenarios with complex height distribution.}
    \label{fig:visz}
\end{figure}

\begin{figure}[h]
    \centering
    \includegraphics[width=1\linewidth]{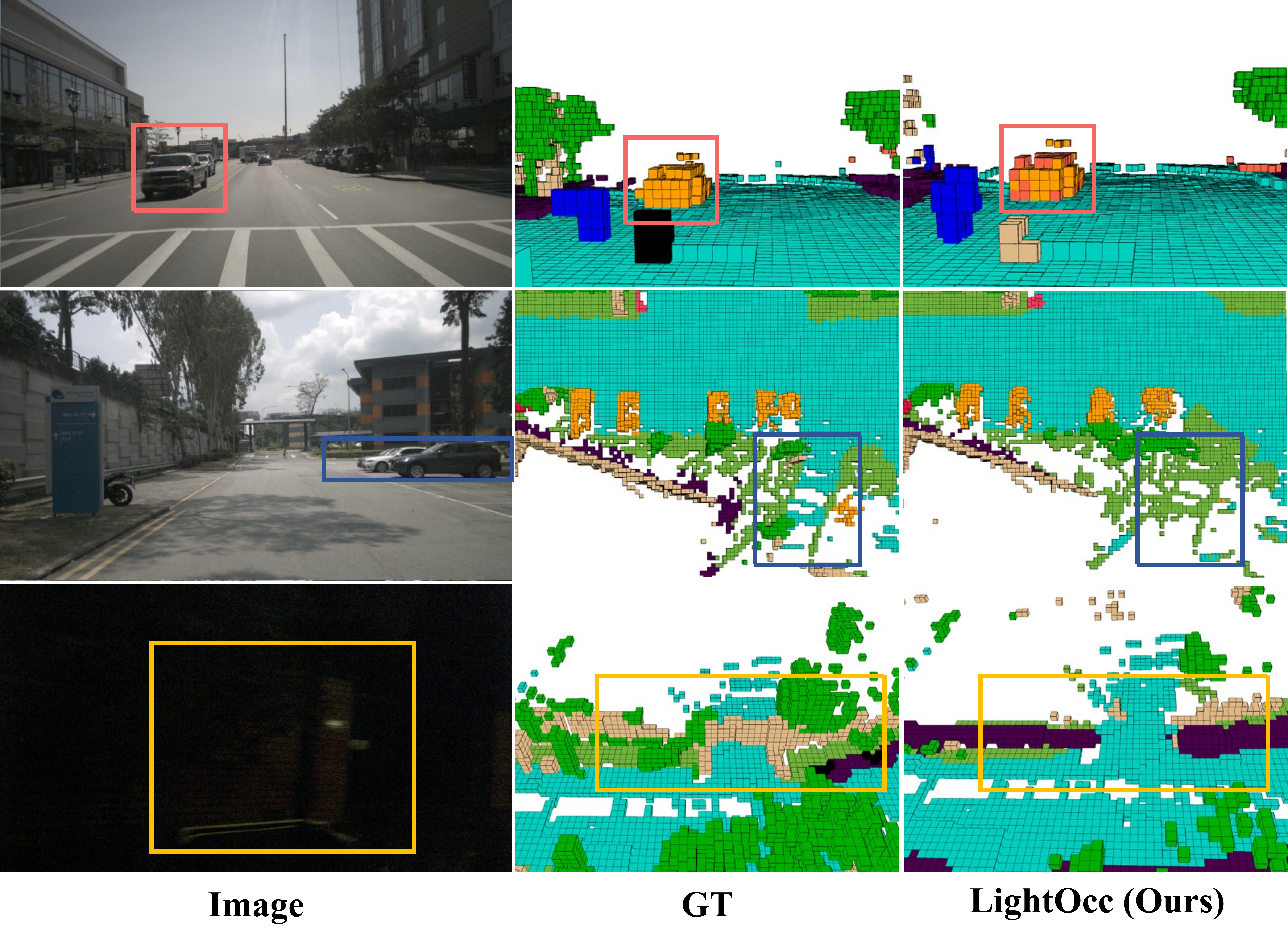}
    \caption{Visualization of Failure Cases predicted by LightOcc on Occ3D-nuScenes dataset.}
    \label{fig:vis_fail}
\end{figure}
 
\subsubsection{Edge-aware Spatial Embedding and Geometric Supervision}

We further conduct the ablation study on Edge-aware Spatial Embedding and Geometric Supervision as shown in Tab.~\ref{tab:edge}. To obtain more reliable edge information, we conduct experiments using LightOcc-S model, which adopts explicit depth supervision. It can be found that Edge Embedding $\mathbf{E}_{U}^{Edge}$ can effectively represent the edge information of the perception space, thereby further improving the accuracy of occupancy prediction. Besides, compared with Sobel operators implemented by 3D convolution, using gradients to represent edges can achieve higher efficiency. Geometric Supervision applied on $\mathbf{E}_{U}$ and $\mathbf{E}_{U}^{Edge}$ can effectively enhance their ability to store the entire spatial information, bringing improvements on mIoU without affecting the efficiency of the model.

We also evaluate IoU of different models. The results show that Edge-aware Spatial Embedding and Geometric Supervision can each improve the accuracy for pure geometric occupancy. The enhancement of spatial embeddings' capability to represent geometric information also consistently improves the model's spatial distribution predictions for objects of different semantic categories, thereby contributing to the final mIoU metric.

\subsubsection{BEV-CutMix}
To avoid introducing incorrect occlusion relationships in the generated scenes, BEV-CutMix cuts the BEV features into quadrants from the center, instead of cutting from random positions. As shown in Tab.~\ref{tab:supp_bev_cutmix}, BEV-CutMix can effectively enhance data diversity and improve the generalization performance of the model. Among different configurations, the setup that crops the scene into two patches along the Y-axis (road direction) yields more realistic synthesized scenes and increases mIoU by 0.66\%. However, the incorrect occlusion relationships introduced by random cutting and mixing will reduce the performance gain by 0.34\% mIoU.

\subsection{Visualization}

We qualitatively compare LightOcc with FlashOcc~\cite{yu2023flashocc} and COTR~\cite{ma2024cotr}, and show their occupancy predictions in Fig.~\ref{fig:visualization}. 
It can be found that the objects' height occupancy predicted by LightOcc is more precise than that predicted by FlashOcc. Specifically, the street lamp and traffic light have more complete structures in our prediction, while they are truncated in the prediction of FlashOcc.
Compared with FlashOCC, COTR has a clear advantage in predicting height structures, benefiting from its compact voxel representation that accurately captures height information. However, the prediction accuracy of COTR still lags behind LightOCC, and according to the results in Tab.~\ref{tab:efficiency}, its FPS is only about 1/3 that of LightOCC.
These visual results demonstrate that LightOcc exhibits superior perception capabilities and robustness in challenging or unconventional driving scenes.

For complex scenes such as tunnels, overpasses, and aerial obstacles, voxels are not all concentrated near the ground, which poses greater challenges for BEV-based methods. We visualized the prediction results of LightOCC in these scenes and present them in Fig~\ref{fig:visz}. It can be observed that LightOCC also achieves strong prediction capability in these scenes, demonstrating that the lightweight spatial embedding is capable of effectively extracting and compressing complex height distribution.

However, the current LightOcc still has several limitations, which introduce discrepancies in its prediction outputs, as illustrated in Fig.~\ref{fig:vis_fail}. Typical examples include misclassification of semantically similar classes, inaccurate predictions for distant regions due to insufficient visual information, and inaccurate modeling of roadside structures under low-light conditions. These challenges represent key directions for our future research and improvement.

\section{Conclusion}

In this paper, we propose LightOcc, a pure 2D convolution model for precise 3D occupancy prediction. The key contribution of LightOcc lies in decoupling the original voxel representation into BEV features and Lightweight Spatial Embeddings, which compactly encode sparse yet essential height information. After obtaining Single-Channel Occupancy by sampling multi-view depth distribution, Spatial-to-Channel mechanism then utilizes 2D convolutions to extract embeddings from different views. Next, Lightweight Cross-View Interaction combines the explicit and implicit height information from these embeddings and obtains the optimal supplement to BEV features. In addition, we extract Edge-aware Spatial Embedding and apply Geometric Supervision to Spatial Embeddings to extract more complete spatial information. BEV-CutMix increases the data diversity without introducing incorrect occlusion relationships. Extensive experiments indicate that LightOcc achieves state-of-the-art performance in 3D occupancy prediction while maintaining the high efficiency of BEV-based methods, making it highly suitable for deployment in autonomous driving applications.

\section*{Data Availability Statements}

The databases used in this manuscript are deposited in publicly available repositories. Occ3D-nuScenes~\cite{tian2024occ3d} can be obtained following the instruction in  \url{https://github.com/Tsinghua-MARS-Lab/Occ3D}. SemanticKITTI~\cite{behley2019semantickitti} can be obtained in \url{https://semantic-kitti.org/dataset.html}. SurroundOcc~\cite{wei2023surroundocc} can be obtained following the instructions in \url{https://github.com/weiyithu/SurroundOcc}. OpenOccupancy~\cite{wang2023openoccupancy} can be obtained following the instructions in\url{https://github.com/JeffWang987/OpenOccupancy}.






\bmhead{Acknowledgements}

This research was supported by Zhejiang Provincial Natural Science Foundation of China under Grant No. LD24F020016, National Natural Science Foundation of China under Grant No. 62506111, the China Postdoctoral Science Foundation under Grant No. 2025M781468, and the Postdoctoral Fellowship Program of CPSF under Grant No. GZC20251098.

\bibliography{sn-bibliography}

\end{document}